\documentclass{article} 
\usepackage{iclr2026_conference,times}

\usepackage{amsmath,amsfonts,bm}

\def\eqref#1{equation~\ref{#1}}

\def\1{\bm{1}}

\DeclareMathAlphabet{\mathsfit}{\encodingdefault}{\sfdefault}{m}{sl}
\SetMathAlphabet{\mathsfit}{bold}{\encodingdefault}{\sfdefault}{bx}{n}

\usepackage{hyperref}
\usepackage{url}
\usepackage{booktabs}
\usepackage{multirow}
\usepackage{colortbl}
\usepackage[normalem]{ulem}
\usepackage[most]{tcolorbox}
\usepackage{bbding}
\useunder{\uline}{\ul}{}

\title{\textbf{PAGE}: Towards \textbf{P}r\textbf{a}ctical Human-level \textbf{G}aze Target \textbf{E}stimation}

\author{%
    Zhoutong Ye$^{1*}$ \And
    Chengwen Zhang$^{1*}$ \And
    Zhaibin Cui$^1$ \And
    Mingze Sun$^1$ \AND
    Jiaqi Liu$^1$ \And
    Xiangwu Li$^2$ \And    
    Qingyang Wan$^1$ \And
    Chang Liu$^1$ \AND
    Xutong Wang$^1$ \And
    Huan-ang Gao$^1$ \And
    Yu Mei$^1$ \And
    Chun Yu$^{1\dagger}$ \And
    Yuanchun Shi$^1$ \AND
    \fontsize{9pt}{10.8pt}\selectfont
    ~~~~~~~~~~$^1$Tsinghua University $^2$Jinan University $^\dagger$Corresponding Author $^*$Equal Contribution \AND
    \fontsize{8pt}{10.8pt}\selectfont
    ~~~~~~~~~~~~~~~~~~~~~~~~~~~~~~~~~~~~~~~~~~~~~~~~~~~~\texttt{\{yezt24, zcw25\}@mails.tsinghua.edu.cn}
}

\newcommand{\Name}{PaGE}

\iclrfinalcopy 
\begin{document}

\maketitle

\begin{abstract}
Gaze target estimation, the task of predicting where a person is looking in a scene, is crucial to understanding human attention and intent. It is a challenging task that combines high-level understanding of global scene semantics and precise spatial reasoning using human appearance (e.g. pose, eye orientation). As a result, human-level performance remains elusive for existing models, limiting their practical application. To this end, we propose \Name{} (\textbf{P}r\textbf{a}ctical \textbf{G}aze \textbf{E}stimator), a gaze estimation model that explicitly models the complex interaction between scene and head features. Using a \Name{} model with a large ViT-H+ backbone as the teacher, we further distill student models with lighter backbones on a much larger and more diverse unlabeled dataset. The architectural improvements and novel training recipe allow \Name{} to achieve state-of-the-art performance on several gaze estimation tasks, outperforming humans in 7 out of 9 metrics while reducing the human-AI gap by at least 60\% in the remaining 2. The distilled student models retain most of the teacher's performance while being lightweight enough for practical deployment on robots and consumer devices. The code and model checkpoints are available at \href{https://PaGE-26.github.io}{https://PaGE-26.github.io}.
\end{abstract}

\section{Introduction}
\label{sec:intro}

Gaze is one of the most important non-verbal social cues. It provides valuable insight into a person's attention and intent, as well as the dynamics of social interactions. It is a key component of socially-aware interactive systems like MLLM agents and robots. Humans perform accurate \textit{gaze following} (i.e., identifying the gaze target of another person in a scene) naturally, yet it is challenging to replicate this capability with vision models. This can be attributed to the inherent complexity of gaze following—it requires a combination of scene understanding and accurate spatial reasoning using human appearance cues (e.g., pose, eye orientation). Therefore, existing models perform substantially worse than humans, limiting their practical application in fields like HCI and robotics.

In this work, we propose \Name{} (Figure \ref{fig:architecture}), \textbf{the first gaze estimation model with human-level performance}. On GazeFollow, VideoAttentionTarget and ChildPlay, \Name{} outperforms humans on 7 out of 9 metrics while closing the current human-AI gap by at least 60\% for the remaining two. Distilled versions of \Name{} retain SOTA performance while being lightweight enough for \textbf{real-time gaze following on robots and many consumer devices}. 

The strong results come from a combination of (1) a \textbf{novel model architecture} designed to explicitly model feature interaction between the scene and head branches, and (2) \textbf{improvements to the training recipe}, an area underexplored in previous work. Specifically, we propose the Scene-head Interaction Module (SIM), a novel gaze decoder component that uses cross attention between the scene and head branches to explicitly model inter-branch feature interaction in a ViT-native manner. This affords \Name{} the spatial reasoning capability needed to pinpoint the gaze target. For training, we adopt a new two-stage approach. We first train the decoder only, with the backbone frozen. We then finetune the entire model, backbone included, to further adapt the model to gaze prediction tasks. To build strong lightweight models, we further propose a token-level feature distillation procedure that trains student models using a \Name{} ViT-H+ teacher. To ensure that the student learns a generalizable feature representation for gaze estimation, we leverage large-scale image data without gaze annotation for distillation as a remedy to the scarcity of labeled data. This results in a line of SotA lightweight \Name{} models with $\sim10\%$ of the teacher's FLOPs that are practical to deploy.

Apart from proposing \Name{}, we also dissect how each component of our architecture and training recipe impacts performance, providing insight for designing future models. We further demonstrate \Name{}'s versatility by adapting it to fine-grained perception of gaze on screens, as well as discussing its potential as a universal visual attention estimator beyond human gaze and realistic images.

\vspace{-0.1cm}
\section{Related Work}

Gaze following is a complex and demanding task requiring a combination of global scene understanding and fine-grained human appearance perception capabilities. One rather intuitive approach is to dedicate a branch to each capability before fusing the features for a final prediction. In addition to the main scene branch, such \textit{multi-branch} models may also include branches for a head crop~\citep{recasens2015gazefollow, chong2020vat, miao2023patchlevel}, depth~\citep{fang2021dualattn, bao2022escnet, jin2022depthaware, gupta2022modular}, and pose~\citep{bao2022escnet, jin2022depthaware, gupta2022modular}.

Another approach emerged as general-purpose vision backbones (e.g., CLIP, DINO) demonstrated state-of-the-art performance in dense vision tasks like depth estimation and segmentation~\citep{simeoni2025dinov3}. In this line of work~\citep{tafasca2024sharingan, song2024vitgaze, ryan2025gaze-lle}, the combination of capabilities is packed into a single large-scale pre-trained backbone. The focus shifts to designing effective gaze decoding strategies. \citet{tafasca2024sharingan} used a DPT decoder conditioned by head features. \citet{song2024vitgaze}'s decoder leverages both patch features and attention maps produced by the ViT encoder. Gaze-LLE~\citep{ryan2025gaze-lle} uses a ViT stack to decode the heatmap from DINOv2 patch features, injecting head position by adding a learnable head prompt to relevant patch tokens. AnyGaze~\citep{cao2026anygaze} explores text prompts as an alternative to head bounding boxes by leveraging an additional pre-trained text encoder. This line of work has led to more flexible, better-performing models than previous multi-branch ones, but is still far from reaching human performance. However, in contrast to recent architectural progress, few have attempted to improve the training recipe of gaze estimation models built upon general-purpose backbones.

Recent progress in gaze target estimation has largely been \textit{model-driven}. Datasets have remained more or less the same, with most work using GazeFollow~\citep{recasens2015gazefollow} and VideoAttentionTarget (VAT)~\citep{chong2020vat}. Some recent work also uses ChildPlay~\citep{tafasca2023childplay} and GOO~\citep{tomas2021goo}. While other datasets have been proposed~\citep{hu2023gfie}, they have yet to gain traction. An obstacle to \textit{data-driven} progress is the tension between accurate annotation and diversity. Datasets like GazeFollow cover diverse scenes sourced from the web but rely on somewhat unreliable human annotations. Meanwhile, datasets with accurate ground truth must be collected in controlled environments under strict procedures~\citep{hu2023gfie, tomas2021goo}.

In this work, we combine the strengths of multi-branch models and the DINOv3~\citep{simeoni2025dinov3} backbone to create \Name{}. Our core architectural contribution is a ViT-native solution for explicit feature interaction between the scene and head branches. We also innovate in how we train gaze models built upon general-purpose backbones, an underexplored area, by introducing SFT and distillation. We further demonstrate that \Name{} is limited by data and has untapped model capacity that can be readily adapted to challenging tasks if trained on proper data.
\section{\Name{}}

\subsection{Problem Definition}
\label{sec:method-problem-def}

2D third-person gaze target estimation can be formally defined as follows: given an RGB image $x_{img} \in \mathbb{R}^{3 \times H_{in} \times W_{in}}$, which must contain at least one person, and a head bounding box $x_{bbox} \in \mathbb{R}^{4}$, the model predicts a heatmap $H \in \mathbb{R}^{H_{out} \times W_{out}}$ where $H_{ij} \in [0, 1]$. The model also predicts $P_{in} \in [0, 1]$, the probability that the person's gaze target is inside the image. An alternative detection-based formulation~\citep{tu2022end2end, tu2023joint, tonini2023objectaware} is discussed in Appendix \ref{sec:appendix-detection-based-methods}.

\subsection{Building \Name{}}
\label{sec:method-arch}

\begin{figure}[h!]
  \centering
  \includegraphics[width=0.95\linewidth]{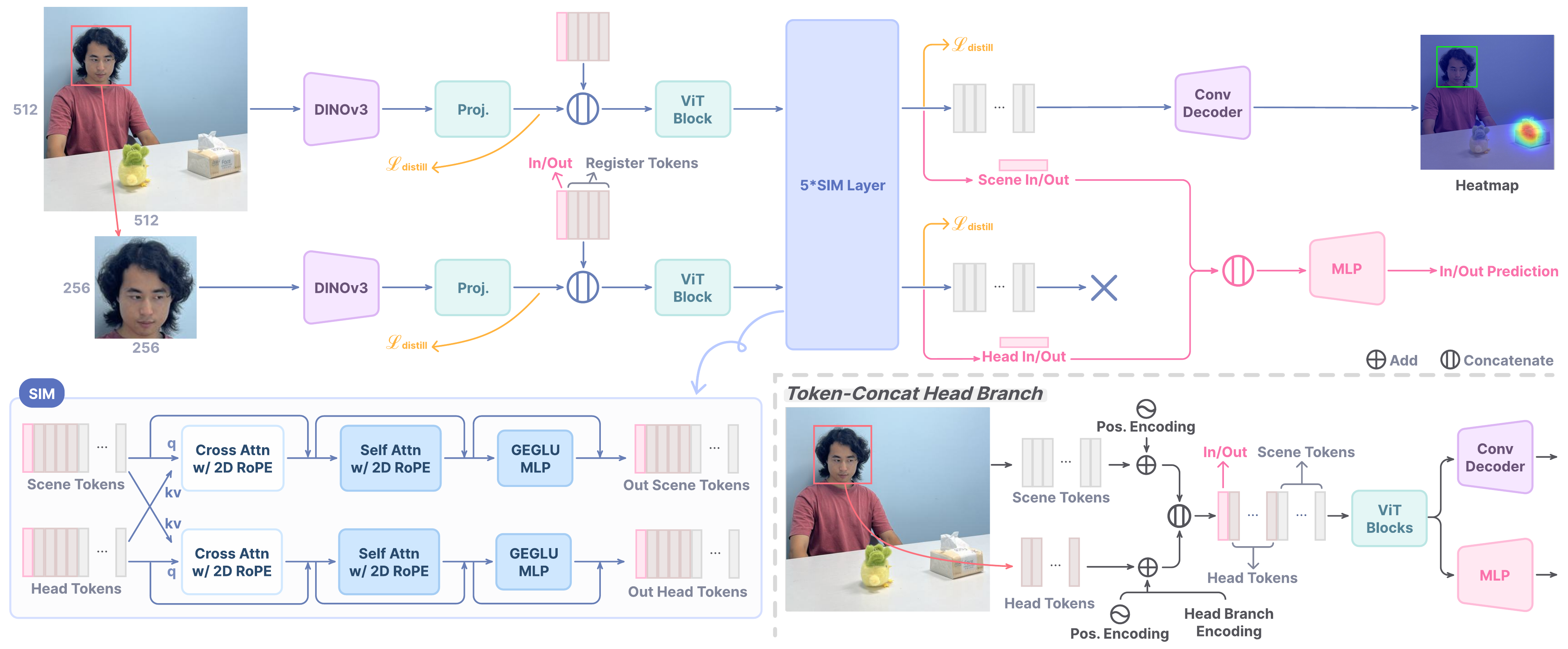}
  \caption{Architecture of \Name{}. Bottom left: our SIM layer. Bottom right: The \textit{token-concat} head branch baseline. We also mark where the $\mathcal{L}_{distill}$ loss is applied during distillation.}
  \label{fig:architecture}
\end{figure}

Figure \ref{fig:architecture} provides an overview of the \Name{} architecture. In this section, we walk through the design choices we made for \Name{}. We start from Gaze-LLE~\citep{ryan2025gaze-lle}, the previous SOTA model, and introduce the changes we made to the backbone, training recipe, and decoder architecture one by one. Gaze-LLE is chosen as the starting point because of its effectiveness and simplicity among prior work. The performance after each change is shown in Table \ref{tab:method-main}.

\begin{table}[htbp]
\scriptsize
\centering
\setlength{\tabcolsep}{5.5pt}
\begin{tabular}{lccccccccc}
\toprule
& \multicolumn{3}{c}{GazeFollow} & \multicolumn{3}{c}{VideoAttentionTarget} & \multicolumn{3}{c}{ChildPlay}  \\
\multirow{-2}{*}{}                               & AUC$\uparrow$      & Avg L2$\downarrow$   & Min L2$\downarrow$   & AUC$\uparrow$        & L2$\downarrow$        & AP{\tiny in/out}$\uparrow$    & AUC$\uparrow$    & L2$\downarrow$     & AP{\tiny in/out}$\uparrow$ \\ \midrule
Gaze-LLE ViT-B$^\dagger$              & 0.9543   & 0.1104   & 0.0492    & 0.9324     & 0.1074    & 0.8970          & 0.9478   & 0.1051 & 0.9920         \\
\rowcolor[HTML]{F6F1FE} 
\cellcolor[HTML]{F6F1FE}Simplify Training Recipe & 0.9552   & 0.1074   & 0.0482   & 0.9370     & 0.1078    & 0.8926          & 0.9499 & 0.1103 & 0.9907       \\
\rowcolor[HTML]{FEF1F1} 
\cellcolor[HTML]{FEF1F1}DINOv2 → DINOv3          & 0.9554   & 0.1044   & 0.0450   & 0.9383     & 0.1118    & 0.9053          & 0.9496 & 0.1062 & 0.9927       \\
\rowcolor[HTML]{F0FBEF} 
\cellcolor[HTML]{F0FBEF}Token-concat Head Branch & 0.9581   & 0.1003   & 0.0439   & 0.9462     & 0.1000    & 0.9138          & 0.9585 & 0.0955 & 0.9942       \\
\rowcolor[HTML]{F0FBEF} 
\cellcolor[HTML]{F0FBEF}3 Layers → 5 Layers      & 0.9583   & 0.0985   & 0.0419   & 0.9516     & 0.0928    & 0.9150          & 0.9604 & 0.0907 & 0.9949       \\
\rowcolor[HTML]{F0FBEF} 
\cellcolor[HTML]{F0FBEF}DINO Feature Dropout     & 0.9591   & 0.0979   & 0.0412   & 0.9502     & 0.0912    & 0.9235          & 0.9596 & 0.0911 & 0.9945       \\
\rowcolor[HTML]{F0FBEF} 
\cellcolor[HTML]{F0FBEF}SIM Head Branch          & 0.9596   & 0.0961   & 0.0392   & 0.9543     & 0.0911    & 0.9304          & 0.9618 & 0.0905 & 0.9949       \\
\rowcolor[HTML]{FEFFF0} 
\cellcolor[HTML]{FEFFF0}Register Tokens          & 0.9595   & 0.0968   & 0.0403   & 0.9544     & 0.0878    & 0.9242          & 0.9608 & 0.0868 & 0.9931       \\
\rowcolor[HTML]{FEFFF0} 
\cellcolor[HTML]{FEFFF0}GEGLU MLP                & 0.9596   & 0.0957   & 0.0399   & 0.9546     & 0.0858    & 0.9277          & 0.9625 & 0.0873 & 0.9944       \\
\rowcolor[HTML]{FEFFF0} 
\cellcolor[HTML]{FEFFF0}2D RoPE                  & 0.9599   & 0.0939   & 0.0380   & 0.9557     & 0.0886    & 0.9421          & 0.9652 & 0.0883 & 0.9961       \\
\rowcolor[HTML]{F0F4FF} 
\cellcolor[HTML]{F0F4FF}Finetune (\Name{} ViT-B) & 0.9625   & 0.0892   & 0.0345   & 0.9618     & 0.0844    & 0.9491          & 0.9690 & 0.0810 & 0.9971       \\ \midrule
Gaze-LLE ViT-L$^\dagger$                         & 0.9569   & 0.1058   & 0.0455   & 0.9371     & 0.1035    & 0.9014           & 0.9505  & 0.1010  & 0.9926        \\
\Name{} ViT-L                                    & 0.9648   & 0.0823   & 0.0306   & 0.9690     & 0.0736    & 0.9460          & 0.9728 & 0.0724 & 0.9957       \\ \midrule
Human$^\ddagger$                                 & 0.924    & 0.096    & 0.040    & 0.921      & 0.051     & 0.925           & 0.911  & 0.048  & 0.993        \\
\Name{} ViT-H+                                   & 0.9659   & 0.0804   & 0.0288   & 0.9719     & 0.0643    & 0.9509          & 0.9746 & 0.0687 & 0.9954       \\ \bottomrule
\end{tabular}
\caption{The road to \Name{}. We track the impact of each modification we made. $^\dagger$Reevaluated with official checkpoints (the ``inout'' version for GazeFollow and VAT, the ``inout\_childplay'' version for ChildPlay) and code. $^\ddagger$Human baseline reported by dataset authors~\citep{recasens2015gazefollow, chong2020vat, tafasca2023childplay}, only three decimal digits are available.}
\label{tab:method-main}
\end{table}

\subsubsection{Simplified Training Recipe}
\label{sec:method-training}

Prior work on gaze estimation trains the model on GazeFollow first before finetuning it on VAT and ChildPlay~\citep{tafasca2024sharingan, ryan2025gaze-lle}. This approach is unnecessarily complex and difficult to integrate with our finetuning (Section \ref{sec:method-sft}) and distillation (Section \ref{sec:method-distillation}) steps. We take a simplified approach instead, training the model on a combined dataset consisting of images from GazeFollow, VAT, and ChildPlay. Specifically, we use the full train split of GazeFollow, sample 1 out of every 3 VAT frames, and 1 out of every 6 ChildPlay frames. We sample VAT and ChildPlay at a reduced rate because adjacent video frames are highly similar (see Appendix \ref{sec:appendix-train-set-ablation} for ablation). An added benefit is that this approach produces a single set of model weights instead of one checkpoint for each dataset, simplifying evaluation and downstream use. We use the standard multitask loss for gaze target estimation: $\mathcal{L} = \mathcal{L}_{heatmap} + \lambda \mathcal{L}_{in/out}$, where $\mathcal{L}_{heatmap}$ and $\mathcal{L}_{in/out}$ denote BCE Loss. We empirically choose $\lambda=0.01$ for all experiments. The rest of the training recipe is detailed in Appendix \ref{sec:appendix-training}. We use the checkpoint from the final epoch for evaluation, finetuning and distillation. As shown in Table \ref{tab:method-main}, the new, simpler training recipe does not impact the results too much. \textbf{All remaining experiments in Section \ref{sec:method-arch} are done with this recipe, and we refer to this train set as the \textit{labeled train set} from now on.}

\subsubsection{DINOv3 Backbone}

We replace DINOv2~\citep{oquab2024dinov2} with DINOv3~\citep{simeoni2025dinov3}, a newer and stronger backbone. Since the patch size of DINOv3 is 16 instead of DINOv2's 14, we increase the input resolution from $448 \times 448$ to $512 \times 512$ to keep the number of patches the same. This simple drop-in replacement results in a modest improvement (Table \ref{tab:method-main}). For more substantial gains, we would need to redesign the decoder architecture. \textbf{We use DINOv3 and $512\times512$ input from now on}. See Appendix \ref{sec:appendix-backbone-ablation} for experiments with alternative backbones (e.g., CLIP, TIPSv2, etc.).

\subsubsection{We DO Need A Head Branch}
\label{sec:method-head-branch}

\citet{ryan2025gaze-lle} found that adding a head branch to Gaze-LLE does not bring significant improvement and hypothesized that DINOv2 extracts sufficient head features from the scene image alone. However, they only explored a flawed implementation where the scene and head feature maps extracted by DINO are directly concatenated along the \textit{channel} dimension. This creates a \textbf{spatial misalignment}. For example, features from the top left of the head crop would be concatenated with features from the top left of the scene. While prior models using convnets~\citep{chong2020vat, miao2023patchlevel} also took this suboptimal approach, the attention mechanism in ViTs allows for more flexibility regarding branch fusion. \textbf{In this section, we show that a well-designed ViT-native head branch, central to our architectural contribution, substantially improves gaze estimation}.

We first consider a head branch architecture that requires the least change (Figure \ref{fig:architecture}, bottom right). Specifically, we extract patch tokens from a $256\times256$ head crop with DINOv3, which are then concatenated with scene tokens along the \textit{length} dimension. We then add a learnable \textit{head branch encoding} to all head tokens to distinguish them from scene tokens. Finally, we pass the tokens through a ViT stack identical to Gaze-LLE's. This \textit{token-concat} head branch significantly improves all metrics (Table \ref{tab:method-main}), suggesting that \textbf{a head branch is indeed needed in our quest for human-level gaze following}. We also find that the head branch helps with \textit{scaling}. While \citet{ryan2025gaze-lle} did not find any upside in stacking more than 3 decoder layers, increasing the number of decoder layers from 3 to 5 is effective for the token-concat model. \textbf{We use 5 decoder layers from now on.} We then apply channel dropout with $p=0.1$ on the DINOv3 feature maps during training. Interestingly, this simple change leads to measurable performance gains, suggesting that DINOv3 could cause our larger, deeper decoder to overfit when used off-the-shelf. \textbf{Channel dropout on backbone features is applied to all models from this point onward}.

Despite its effectiveness, the token-concat approach still has room for improvement. It forces scene features and head features into the same feature space by concatenating the branches into the same token sequence. This motivated us to design the Scene-head Interaction Module (SIM, Figure \ref{fig:architecture}, bottom left), which keeps scene and head features in separate branches and explicitly models inter-branch feature interaction with cross attention instead. After cross attention with the other branch, both branches are passed through a standard ViT block consisting of self attention and an FFN. A stack of SIMs could then model complex interactions between scene and head features. Finally, considering that we have two separate branches now, we append an In/Out token to both branches and concatenate them to create the input for the In/Out prediction head. The results in Table \ref{tab:method-main} show that decoupling the two branches is effective. Results on alternative SIM architectures are in Appendix \ref{sec:appendix-layer-ablation}. \textbf{All decoders from now on use SIM layers instead of vanilla ViT blocks.}

\subsubsection{Modernizing the ViT Decoder}

Gaze-LLE uses vanilla ViT blocks as its decoder. Although we replaced ViT blocks with SIM modules in Section \ref{sec:method-head-branch}, SIM itself is still made up of plain ViT components. In this section, we introduce common modern ViT components. We first add 4 register tokens to the scene and head branches. We then replace the standard GELU MLPs with GLU-style MLPs. We tried SwiGLU and GEGLU, and GEGLU performed better (Appendix \ref{sec:appendix-glu}). \textbf{We keep these changes in SIM.}

Finally, we replace the 2D sinusoidal positional encoding with 2D RoPE, which is also used in DINOv3. For self attention, the standard axial 2D RoPE implementation proposed by \citet{heo2024rotary} is used. For cross attention, the scene branch uses standard RoPE, while the head branch calculates RoPE based on the scene branch coordinates of the head crop. See Appendix \ref{sec:appendix-rope-formulation} for the exact formulation. An added benefit of 2D RoPE is that it \textbf{implicitly encodes head position}, as both branches operate in the same coordinate system. Therefore, we remove the \textit{learnable head prompt} introduced by Gaze-LLE. As shown in Table \ref{tab:method-main}, 2D RoPE is especially effective on GazeFollow, as well as predicting whether the gaze target is in-frame (AP{\tiny in/out}). \textbf{We have now arrived at the final architecture of \Name{}, a strong and modern gaze target estimation model. It will be used in the rest of this paper.} See Appendix \ref{sec:appendix-more-decoder-ablation} for more extensive ablation studies of the architecture.

\subsubsection{Finetuning the Backbone}
\label{sec:method-sft}

Until now, we have kept the DINOv3 backbone frozen. This is because making the backbone trainable from the start actually degrades performance (see Section \ref{sec:exp-ft-distill-ablation} for experiments). However, a short supervised finetuning (SFT) regime with the backbone unfrozen \textit{after} the decoder is trained does bring concrete improvements. Specifically, we use the same labeled train set, and train the scene and head branch backbones separately to allow for branch-specific feature adaptation, although both backbones start from DINOv3 weights. The detailed finetuning procedures are in Appendix \ref{sec:appendix-training}. As shown in Table \ref{tab:method-main}, SFT significantly improved performance on all metrics. \textbf{The finetuned ViT-H+ variant, in particular, achieves human-level performance, outperforming humans on 7 out of 9 metrics. It is now ready to be used as the teacher model for distillation.}

\subsection{Feature Distillation with Unlabeled Images}
\label{sec:method-distillation}

In previous sections, we achieve strong performance when a heavy ViT-H+ (840M Params) backbone is used. However, the results of smaller variants (e.g., \Name{} ViT-B), while still SOTA, lag behind those of \Name{} ViT-H+ and humans. Therefore, in this section, we explore how to \textbf{transfer the superior capabilities of \Name{} ViT-H+ to smaller variants through feature distillation.}

\subsubsection{Dataset Curation}
\label{sec:method-distillation-data}

Knowledge distillation works well when the dataset is large and diverse~\citep{frank2025kddataset}. Considering that labeled gaze datasets are limited in both of these regards, we turn to public image datasets without gaze annotation instead. Specifically, we combine MPII~\citep{andriluka2014mpii}, a human pose dataset with high quality images of people, and OpenImages V7~\citep{krasin2017openimages}, a much larger dataset with object annotations. The resulting distillation set consists of all MPII training samples and all OpenImages V7 training samples containing objects labeled ``\textit{person}'', ``\textit{man}'', ``\textit{woman}'', ``\textit{boy}'', and ``\textit{girl}''. We do not use MS COCO~\citep{lin2014coco}, a major source of GazeFollow test images, to avoid data leakage. We then follow \citet{ryan2025gaze-lle} and use a YOLOv5 head detector to provide head bounding boxes for people inside each image. For each image, we keep a maximum of 3 heads to ensure scene diversity. The final distillation set consists of $1.17M$ heads from $608k$ images, significantly more than the $200k$ samples in our labeled train set. A simple experiment validating the quality of the unlabeled distillation set is in Appendix \ref{sec:appendix-distillation-set-quality}.

\subsubsection{Distillation Objective}
\label{sec:method-distillation-objective}

We conduct token-level feature distillation to train lightweight student models that retain the \Name{} ViT-H+ teacher's strong performance. The main objective is to align the student's final-layer (i.e., after the last SIM layer) token representation to the teacher's. Formally, for an input pair $\{x_{img}, x_{bbox}\}$, let $\mathbf{z}^{s}, \mathbf{z}^{t} \in \mathbb{R}^{d}$ denote the student and teacher tokens, respectively. We optimize a distillation loss that combines element-wise $\ell_1$ loss with a cosine alignment term:
\[
\mathcal{L}_{\mathrm{distill}}
=
\left\|
\mathbf{z}^{s} - \mathbf{z}^{t}
\right\|_{1}
+
\left(
1 -
\frac{
\left\langle \mathbf{z}^{s}, \mathbf{z}^{t} \right\rangle
}{
\left\|\mathbf{z}^{s}\right\|_{2}
\left\|\mathbf{z}^{t}\right\|_{2}
}
\right).
\]
The $\ell_1$ term encourages absolute feature matching, while the cosine term encourages directional alignment. The loss is applied to both scene and head branch tokens. We then add an \textit{auxiliary objective} that aligns the student's \textit{backbone features} to the teacher's right after they are both projected to the decoder dimension (Figure \ref{fig:architecture}). The idea is to prevent representation collapse in the student backbones. We use the same $\mathcal{L}_{distill}$ objective and add the auxiliary loss directly to the main loss. We train both the backbone (initialized with DINOv3 weights) and the \Name{} decoder (random initialization). As in Section \ref{sec:method-sft}, the scene and head backbones are trained separately to allow for branch-specific feature adaptation. Since distillation only teaches the student feature representation, not downstream prediction, we add a short post-distillation finetuning regime on the labeled train set. The exact procedure and hyperparameters are detailed in Appendix \ref{sec:appendix-distillation}. 

\begin{table}[htbp]
\scriptsize
\centering
\setlength{\tabcolsep}{5pt}
\begin{tabular}{lcccccccccc}
\toprule
\multirow{2}{*}{Model}     & \multirow{2}{*}{GFLOPs} & \multicolumn{3}{c}{GazeFollow} & \multicolumn{3}{c}{VideoAttentionTarget} & \multicolumn{3}{c}{ChildPlay} \\
                           &                         & AUC      & Avg L2   & Min L2   & AUC          & L2          & AP          & AUC      & L2       & AP      \\ \midrule
\Name{} ViT-S            & \multirow{2}{*}{96.9}   & 0.9605   & 0.0978   & 0.0411   & 0.9516       & 0.0954      & 0.9218      & 0.9641   & 0.0899   & 0.9961  \\
\Name{} ViT-S Distill    &                         & 0.9643   & 0.0858   & 0.0327   & 0.9640       & 0.0739      & 0.9371      & 0.9701   & 0.0751   & 0.9966  \\ \midrule
\Name{} ViT-S+           & \multirow{2}{*}{115.2}  & 0.9611   & 0.0930   & 0.0370   & 0.9548       & 0.0919      & 0.9302      & 0.9652   & 0.0868   & 0.9963  \\
\Name{} ViT-S+ Distill   &                         & 0.9647   & 0.0861   & 0.0331   & 0.9654       & 0.0740      & 0.9388      & 0.9704   & 0.0746   & 0.9971  \\ \midrule
\Name{} ViT-B            & \multirow{2}{*}{283.1}  & 0.9625   & 0.0892   & 0.0345   & 0.9618       & 0.0844      & 0.9491      & 0.9690   & 0.0810   & 0.9971  \\
\Name{} ViT-B Distill    &                         & 0.9660   & 0.0814   & 0.0295   & 0.9688       & 0.0677      & 0.9450      & 0.9734   & 0.0697   & 0.9969  \\ \midrule
\Name{} ViT-H+ (Teacher) & 2373.6                  & 0.9659   & 0.0804   & 0.0288   & 0.9719       & 0.0643      & 0.9509      & 0.9746   & 0.0687   & 0.9954  \\ \bottomrule
\end{tabular}
\caption{Comparison between \Name{} models trained from scratch and ones distilled from \Name{} ViT-H+ teacher. Distilled models perform significantly better and retain most of the teacher's capability.}
\label{tab:method-distillation-result}
\end{table}

We report the performance of the distilled models in Table \ref{tab:method-distillation-result}. The distilled models performed significantly better than identical models trained from scratch and retained most of the teacher's performance. This proves that feature distillation enables \textit{efficient} human-level gaze target estimation with roughly 5\%-10\% of the teacher's compute burden. Detailed profiling is in Appendix \ref{sec:appendix-sft-distill-profiling}.

\section{Further Experiments and Results}

\subsection{Full Results}
\label{sec:exp-full-results}

\begin{table}[htbp]
    \scriptsize
    \centering
    \begin{tabular}{lccccccccccc} 
        \toprule
                 & \multicolumn{3}{c}{GazeFollow} & \multicolumn{3}{c}{VideoAttentionTarget} & \multicolumn{3}{c}{ChildPlay} \\
         Model   & AUC$\uparrow$ & Avg L2$\downarrow$  & Min L2$\downarrow$ & AUC$\uparrow$ & L2$\downarrow$  & AP{\tiny in/out}$\uparrow$ & AUC$\uparrow$ & L2$\downarrow$  & AP{\tiny in/out}$\uparrow$ \\
         \midrule
         \citet{recasens2015gazefollow}  & 0.878 & 0.190& 0.113 & - & - & - & - & - & - \\
         \citet{chong2018connecting}   & 0.896&  0.187&  0.112 & 0.833 & 0.171 & 0.712 & - & - & - \\ 
         \citet{lian2018believe}  & 0.906&  0.145&  0.081 & - & - & - & - & - & - \\
         \citet{chong2020vat}   & 0.921&  0.137&  0.077& 0.860 &  0.134&  0.853 & - & - & - \\ 
         \citet{fang2021dualattn}  & 0.922&  0.124&  0.067& 0.905&  0.108 &  0.896 & - & - & - \\ 
         \citet{bao2022escnet}   & 0.928&  0.122&  -& 0.885&  0.120 &  0.869 & - & - & - \\ 
         \citet{jin2022depthaware} & 0.920 &  0.118&  0.063 & 0.900 &  0.104 & 0.895 & - & - & -\\ 
         \citet{gupta2022modular}   & 0.943 & 0.114 & 0.056 & 0.914 & 0.110 & 0.879 & 0.919 & 0.113 & 0.983 \\
         \citet{miao2023patchlevel} & 0.934& 0.123& 0.065 & 0.917 & 0.109 & 0.908 & - & - & - \\
         \citet{tafasca2023childplay} & 0.939& 0.122& 0.062 & 0.914 & 0.109 & 0.834 & 0.935 & 0.107 & 0.986 \\
         \citet{tafasca2024sharingan} & 0.944 & 0.113 & 0.057 & - & 0.107 & 0.891 & - & 0.106 & 0.990\\
         \citet{song2024vitgaze}  & 0.949 & 0.105 & 0.047 & 0.938 & 0.102 & 0.905 & - & - & - \\
         \citet{ryan2025gaze-lle}$^*$ & 0.958 & 0.099 & 0.041 & 0.937 & 0.103 & 0.903 & 0.951 & 0.101 & 0.994 \\ 
         \midrule
         Gemini 3.5 Flash & - & 0.128 & 0.067 & - & 0.138 & - & - & 0.111 & - \\ 
         \midrule
         \Name{} ViT-S Distill & 0.964 & 0.086 & 0.033 & 0.964 & 0.074 & 0.937 & 0.970 & 0.075 & 0.997 \\ 
         \Name{} ViT-S+ Distill & 0.965 & 0.086 & 0.033 & 0.965 & 0.074 & 0.939 & 0.970 & 0.075 & \textbf{0.997} \\ 
         \Name{} ViT-B Distill & \textbf{0.966} & {\ul 0.081} & {\ul 0.029} & {\ul 0.969} & {\ul 0.068} & {\ul 0.945} & {\ul 0.973} & {\ul 0.070} & {\ul 0.997}\\ 
         \Name{} ViT-H+ & {\ul 0.966} & \textbf{0.080} & \textbf{0.029} & \textbf{0.972} & \textbf{0.064} & \textbf{0.951} & \textbf{0.975} & \textbf{0.069} & 0.995\\
         \midrule
         Human & 0.924 & 0.096 & 0.040 &0.921 & 0.051 & 0.925 & 0.911  & 0.048  & 0.993 \\
         \bottomrule
    \end{tabular}
    \caption{Comparison between prior art, the \Name{} family, and humans. All four \Name{} models far outperform the previous SOTA, with \Name{} ViT-H+ and \Name{} ViT-B Distill achieving human-level performance. $^*$Gaze-LLE ViT-L, results reported in the original paper.}
    \label{tab:full-results-gf-vat-cp} 
\end{table}

We provide a full comparison with prior work on GazeFollow, VAT and ChildPlay in Table \ref{tab:full-results-gf-vat-cp}. We also include Gemini 3.5 Flash, a SOTA multimodal LLM in our comparison (see Appendix \ref{sec:appendix-gemini-eval} for MLLM evaluation procedure). On GazeFollow, \Name{} exceeded human performance on all three metrics. As each test sample in GazeFollow has around 10 human annotations, our results demonstrate that \Name{} is better than individual humans at aligning with the \textit{overall human consensus}. On VAT, \Name{} outperformed humans on AUC and AP{\tiny in/out}, while reducing the human-AI gap on L2 error by 75\% from 0.052 (Gaze-LLE ViT-L) to 0.013 (\Name{} ViT-H+). Similarly, on ChildPlay, \Name{} outperformed humans on AUC and AP{\tiny in/out} while reducing the human-AI gap on L2 error by 60\% from 0.053 (Gaze-LLE ViT-L) to 0.021 (\Name{} ViT-H+). \textbf{We are the first to push these human-annotated benchmarks, where human performance is a soft ceiling, towards saturation.}

\begin{table}[h!]
\centering
\scriptsize
\begin{tabular}{lllc}
\toprule
\multirow{2}{*}{Model}                       & \multicolumn{2}{c}{GOO-Real}     & \multicolumn{1}{c}{\citet{zhang2025gaze_target_benchmark}}  \\
                  & AUC $\uparrow$ & L2 $\downarrow$ & \multicolumn{1}{c}{Accuracy} \\
\midrule
\citet{chong2020vat}                & 0.670$^*$      & 0.334$^*$       & -                            \\
\citet{miao2023patchlevel}          & 0.869$^*$      & 0.202$^*$       & -                            \\
\citet{ryan2025gaze-lle}$^{\dagger}$ & 0.898$^*$    & 0.175$^*$       & 45.40                        \\
\Name{} ViT-S Distill               & 0.914          & 0.151           & 72.42                        \\
\Name{} ViT-S+ Distill              & {\ul 0.915}    & {\ul 0.149}     & {\ul 72.90}                   \\
\Name{} ViT-B Distill               & 0.914          & 0.153           & \textbf{76.29}                \\
\Name{} ViT-H+ Teacher              & \textbf{0.930} & \textbf{0.140}  & 71.98                        \\
\bottomrule
\end{tabular}
\caption{Results on GOO-Real and \citet{zhang2025gaze_target_benchmark}'s benchmark. On both datasets, \Name{} achieved SOTA without training. $^*$Results reported by \citet{ryan2025gaze-lle}. $^\dagger$Gaze-LLE ViT-L.}
\label{tab:full-results-goo}
\end{table}

\begin{figure}[h!]
  \centering
  \includegraphics[width=0.90\linewidth]{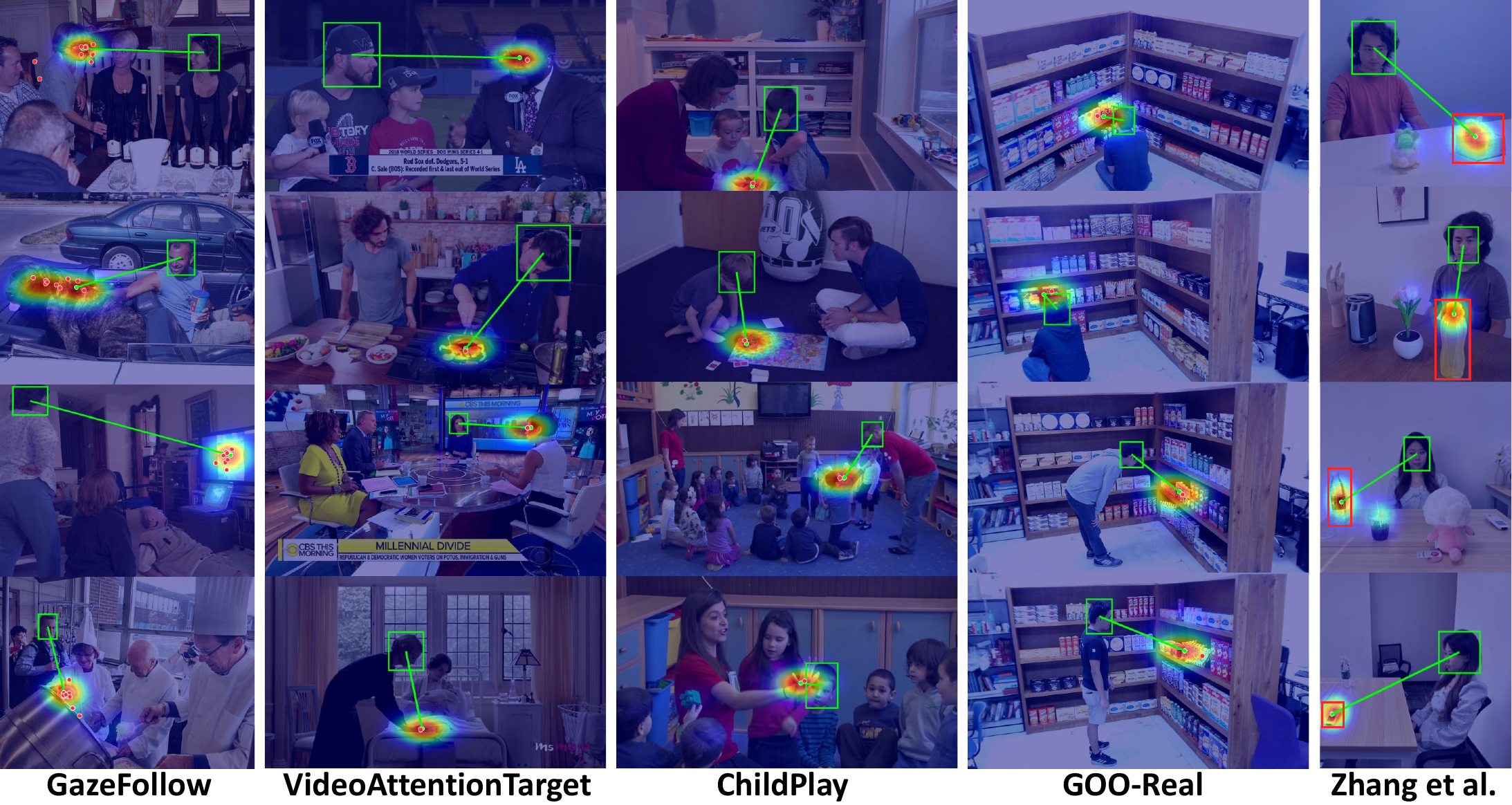}
  \caption{Qualitative results of \Name{} ViT-H+. Red dots and bounding boxes are ground truths.}
  \label{fig:qualitative-main-text}
\end{figure}

To show that \Name{} generalizes across datasets, we evaluate it on GOO-Real~\citep{tomas2021goo} and a benchmark emphasizing head and eye orientation~\citep{zhang2025gaze_target_benchmark} \textbf{without additional training}. The results are in Table \ref{tab:full-results-goo}. The L2 error on GOO-Real is larger than that on other datasets since many images in GOO-Real have the person facing away from the camera. In these cases, humans also struggle to identify the gaze target. However, since \citet{tomas2021goo} did not provide a human baseline, the human-AI gap cannot be measured. Meanwhile, on Zhang et al.'s benchmark, \Name{} achieves SOTA performance. Since that benchmark is challenging for MLLMs~\citep{zhang2025gaze_target_benchmark}, the result underscores \Name{}'s potential in ``Gaze + MLLM'' pipelines like GazeCoT~\citep{ye2026gazecot}. Qualitative results for all 5 datasets are in Figure \ref{fig:qualitative-main-text}.

\subsection{Analysis of Finetuning and Feature Distillation}
\label{sec:exp-ft-distill-ablation}

\begin{table}[h!]
\scriptsize
\centering
\begin{tabular}{lccccccccc}
\toprule
\multirow{2}{*}{Experiment}    & \multicolumn{3}{c}{GazeFollow} & \multicolumn{3}{c}{VideoAttentionTarget} & \multicolumn{3}{c}{ChildPlay} \\
                               & AUC      & Avg L2   & Min L2   & AUC          & L2          & AP          & AUC      & L2       & AP      \\ \midrule
Train Decoder Only, no SFT     & {\ul 0.9599}   & {\ul 0.0939}   & {\ul 0.0380}   & {\ul 0.9557}       & {\ul 0.0886}      & {\ul 0.9421}      & {\ul 0.9652}   & {\ul 0.0883}   & {\ul 0.9961}  \\
Train Full Model (End to End)  & 0.9073   & 0.1881   & 0.1175   & 0.8795       & 0.1733      & 0.8449      & 0.9011   & 0.1632   & 0.9842 \\
Train Decoder, then SFT        & \textbf{0.9625}   & \textbf{0.0892}   & \textbf{0.0345}   & \textbf{0.9618}       & \textbf{0.0844}      & \textbf{0.9491}      & \textbf{0.9690}   & \textbf{0.0810}   & \textbf{0.9971}  \\ \midrule
Distill w/ 100k Samples + SFT  & 0.9639   & 0.0868   & 0.0333   & 0.9649       & 0.0724      & {\ul 0.9463}      & 0.9703   & 0.0742   & 0.9945        \\
Distill w/ 200k Samples + SFT  & 0.9647   & 0.0839   & 0.0316   & 0.9658       & {\ul 0.0718}      & \textbf{0.9484}      & 0.9710   & 0.0728   & 0.9958  \\
Distill w/ 400k Samples + SFT  & {\ul 0.9651}   & {\ul 0.0834}   & {\ul 0.0307}   & {\ul 0.9666}       & 0.0721      & 0.9415      & {\ul 0.9713}   & {\ul 0.0719}   & {\ul 0.9965}        \\
Distill w/ 1.17M Samples + SFT & \textbf{0.9660}   & \textbf{0.0814}   & \textbf{0.0295}   & \textbf{0.9688}       & \textbf{0.0677}      & 0.9450      & \textbf{0.9734}   & \textbf{0.0697}   & \textbf{0.9969}  \\ \bottomrule
\end{tabular}
\caption{Ablation study for SFT and distillation. All experiments are done with \Name{} ViT-B.}
\label{tab:sft-distill-ablation-main}
\end{table}

In this section, we ablate elements of SFT and feature distillation, which are central to our contribution to the training recipe, and analyze their performance impact. First, we show that training the full model only works when we have already trained a decent decoder on a frozen backbone. Training the full model from the start results in very poor performance (Table \ref{tab:sft-distill-ablation-main}). We hypothesize that a randomly initialized decoder unable to make sense of backbone features would create noisy gradients that hinder the backbone from learning useful adaptations on top of DINOv3. Second, we compare student models distilled with different amounts of unlabeled samples. More distillation data consistently leads to better performance. Together, these results validate our novel training recipe, which could also be used by future gaze estimation models. More analysis is in Appendix \ref{sec:appendix-sft-distill-ablation}.

\subsection{You Can Teach an Old Dog New Tricks}
\label{sec:exp-gaze-on-screen}
To demonstrate that \Name{} is limited by existing gaze datasets, we explore the specialized task of tracking gaze on a computer screen~\citep{liu2026aamultiviewmultimodaldataset}. We choose this task because it requires spatial perception and reasoning beyond human capabilities. We use data from 21 users for training, and 3 users for evaluation. We combine the train set used in previous sections (a mix of GazeFollow, VAT and ChildPlay, 200k images) with the new data (20k images) and finetune the \Name{} ViT-B student on this expanded train set. We report results in Table \ref{tab:results-gaze-on-screen}.

\begin{table}[htbp]
\scriptsize
\centering
\begin{tabular}{lccccccccc}
\toprule
\multicolumn{1}{c}{}                   & \multicolumn{3}{c}{GazeFollow}                                                             & \multicolumn{3}{c}{VideoAttentionTarget}                                                        & \multicolumn{3}{c}{\textbf{\citet{liu2026aamultiviewmultimodaldataset}}}                                                               \\
\multicolumn{1}{c}{\multirow{-2}{*}{}} & AUC$\uparrow$                          & Avg L2$\downarrow$                       & Min L2$\downarrow$                       & AUC$\uparrow$                          & L2$\downarrow$                           & AP{\tiny in/out}$\uparrow$ & AUC$\uparrow$                          & L2$\downarrow$                            & AP{\tiny in/out}$\uparrow$ \\ \midrule
SFT w/o Screen Data                    & 0.9660                       & 0.0814                       & 0.0295                       & 0.9688                       & 0.0677                       & 0.9450                            & 0.9562                       & 0.0961                        & 0.9978                            \\
SFT w/   Screen Data                   & 0.9662                       & 0.0806                       & 0.0294                       & 0.9685                       & 0.0690                       & 0.9375                            & 0.9858                       & 0.0524                        & 0.9991                            \\
Performance Impact                  & {\color[HTML]{009901} +0.02\%} & {\color[HTML]{009901} -0.98\%} & {\color[HTML]{009901} -0.34\%} & {\color[HTML]{CB0000} -0.03\%} & {\color[HTML]{CB0000} +1.92\%} & {\color[HTML]{CB0000} -0.79\%}      & {\color[HTML]{009901} +3.10\%} & {\color[HTML]{009901} \textbf{-45.47\%}} & {\color[HTML]{009901} +0.13\%}      \\ \bottomrule
\end{tabular}
\caption{Including gaze-on-screen data when finetuning the student model significantly improves performance on that task (-45\% L2 error) without impacting performance elsewhere too much.}
\label{tab:results-gaze-on-screen}
\end{table}

\begin{figure}[h!]
  \centering
  \includegraphics[width=0.9\linewidth]{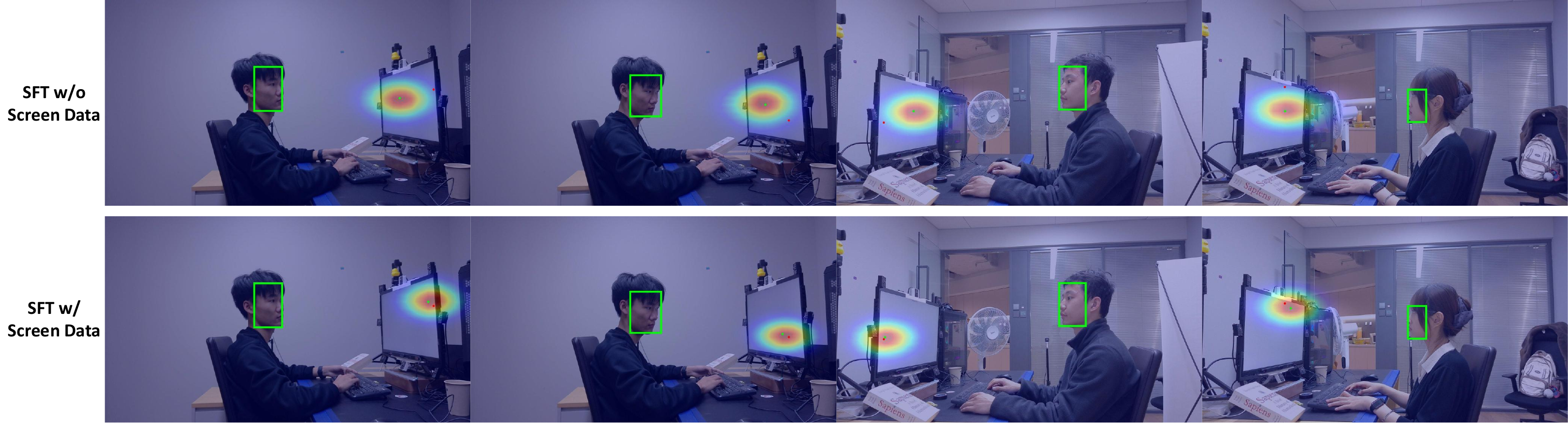}
  \caption{Qualitative results on Liu et al.'s dataset. Red dots are ground truth annotations.}
  \label{fig:qualitative-screen}
\end{figure}

The results show that the added data did not impact GazeFollow and VAT metrics by much, but significantly improved gaze tracking on screens. This is evident in Figure \ref{fig:qualitative-screen}: without the additional data, the model simply guessed that the person was looking at the center of the screen; with additional training, however, the model learned to leverage nuanced details of the person's eye and head. \textbf{This demonstrates that \Name{} still has considerable untapped model capacity. Diverse datasets with accurate ground truth are essential for future data-driven progress in gaze estimation.}

\subsection{Are We Evaluating Human-level Models Fairly?}
\label{sec:exp-eval-bias}


\Name{} nearly saturates existing metrics. At this point, the errors introduced by the evaluation procedure begin to matter. When calculating L2 errors, the standard evaluation procedure of VAT and ChildPlay derives the normalized coordinates of the model's predicted gaze target from a $64\times64$ heatmap, while humans annotate the images at native resolution. This creates room for \textit{discretization errors}. Specifically, open-source evaluation scripts used in previous work~\citep{chong2020vat, ryan2025gaze-lle} calculate the predicted normalized gaze target as {$\hat{g}=(\frac{\hat{x}}{64},\frac{\hat{y}}{64})$}, introducing a shift towards the top left of the image. We correct for this bias and use {$\hat{g}=(\frac{\hat{x}+0.5}{64},\frac{\hat{y}+0.5}{64})$}, and report the results in Table \ref{tab:exp-fix-eval-bias}. It is remarkable that this simple error accounts for 12\%-14\% of the apparent L2 gap between \Name{} and humans. Therefore, future datasets should use unbiased evaluation or higher resolution heatmaps (see Appendix \ref{sec:appendix-512px-heatmap}) for fairness. \textbf{Note that the results reported elsewhere in this paper still use the standard biased protocols to maintain comparability with prior work.}

\begin{table}[h!]
\scriptsize
\centering
\setlength{\tabcolsep}{8pt}
\begin{tabular}{lcccc}
\toprule
\multirow{2}{*}{\begin{tabular}[c]{@{}l@{}}Evaluation\\ Protocol\end{tabular}} & \multicolumn{2}{c}{GazeFollow} & VideoAttentionTarget                  & ChildPlay                             \\
                                                                               & Avg L2         & Min L2        & L2                                    & L2                                    \\ \midrule
Biased                                                                         & 0.0804         & 0.0288        & 0.0643                                & 0.0687                                \\
Unbiased                                                                       & 0.0786         & 0.0277        & 0.0624                                & 0.0662                                \\
Human-AI Gap Reduction                                                         & -              & -             & -14.3\% (0.0133 $\rightarrow$ 0.0114) & -12.1\% (0.0207 $\rightarrow$ 0.0182) \\ \bottomrule
\end{tabular}
\caption{Removing the systematic bias in evaluation. No human-AI gap is reported for GazeFollow because \Name{} ViT-H+ surpassed the human baseline on both GazeFollow L2 metrics.}
\label{tab:exp-fix-eval-bias}
\end{table}

\subsection{Discussion: Is \Name{} a Universal Visual Attention Estimator?}
\label{sec:exp-animal-and-animation}

\begin{figure}[h!]
  \centering
  \includegraphics[width=0.90\linewidth]{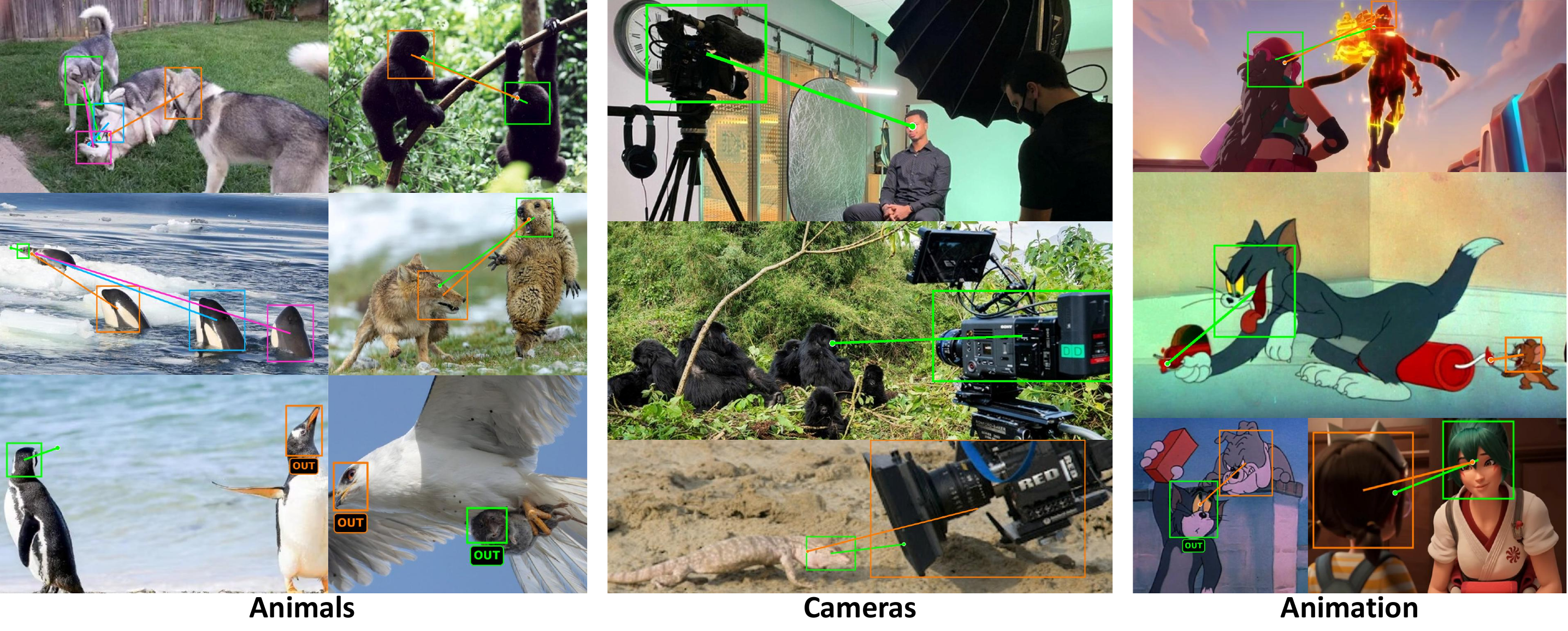}
  \caption{Zero-shot visual attention estimation beyond real-world human gaze. \Name{} can generalize to animals, cameras, and animation. Do you agree with \Name{}'s predictions?}
  \label{fig:discussion-universal-visual-attention}
\end{figure}

\textit{How far can generalization take us?} We are pleasantly surprised to find that \Name{} can estimate the gaze of \textit{animals} to some degree. In fact, for the examples in Figure \ref{fig:discussion-universal-visual-attention}, our human intuition mostly agrees with \Name{}'s predictions across many different species. Even more surprisingly, we found that \Name{} can predict which area a \textit{camera} is filming. Part of this generalizability could plausibly be attributed to the vast train set of DINOv3. In addition to that, we hypothesize that \Name{} has learned a type of \textbf{\textit{universal representation of visual attention}} through training on human gaze data alone. This representation combines global features (e.g., social dynamics, saliency) and local ones (e.g., head and eye orientation, pose), and can partially generalize to animals and cameras. For example, the primary cue in the eagle-and-rodent image is the animals' eye orientation; the gorilla case, despite blurry faces, still has pose and social dynamics; the penguin example offers a strong pose cue (the animal is facing away from the camera). Meanwhile, we hypothesize that \Name{} predicts camera targets primarily through pose and saliency cues.

Another type of generalization concerns \textit{non-realistic images}. To this end, we test \Name{} on animation frames with various styles (Figure \ref{fig:discussion-universal-visual-attention}), which often break the fine-grained local features (e.g., eye features) and strict spatial priors found in realistic images. While \Name{} does encounter some challenges as expected (e.g., the wrong in/out prediction at the bottom-left of the \textit{animation} column shows that \Name{} has limited understanding of animated eyes), it still works reasonably well in many cases. This provides further qualitative evidence that \Name{} has learned a universal, comprehensive representation of visual attention that still works in artificial images that break real-world priors.
\section{Conclusion}

We proposed \Name{}, the first human-level gaze target estimation model. \Name{} achieved SOTA performance and high efficiency through (1) a novel architecture designed to model complex interactions between scene and head features, (2) an improved training recipe that combines multiple datasets, decoder training, and full-scale finetuning, and (3) feature distillation on large-scale unlabeled data. We dissected how each design choice contributed to performance, explored expanding model capability to a new task, and designed fairer evaluation protocols. We ended by discussing \Name{}'s ability to generalize beyond human gaze and realistic images, paving the way for a \textit{GazeAnything} model. We hope our work can turn a new \textit{page} for the field of gaze target estimation and drive the development of better models, datasets, and downstream applications.

\bibliography{iclr2026_conference}

@InProceedings{ryan2025gaze-lle,
    author    = {Ryan, Fiona and Bati, Ajay and Lee, Sangmin and Bolya, Daniel and Hoffman, Judy and Rehg, James M.},
    title     = {Gaze-LLE: Gaze Target Estimation via Large-Scale Learned Encoders},
    booktitle = {Proceedings of the IEEE/CVF Conference on Computer Vision and Pattern Recognition (CVPR)},
    month     = {June},
    year      = {2025},
    pages     = {28874-28884}
}

@inproceedings{ye2026gazecot,
author = {Ye, Zhoutong and Wang, Xutong and Zhang, Chengwen and Zhang, Ruiwen and Sun, Mingze and Li, Qinwei and Yu, Chun and Shi, Yuanchun},
title = {GazeCoT: Unleashing Social Intelligence in Multimodal LLMs With Gaze-Informed Chain-of-Thought Reasoning},
year = {2026},
isbn = {9798400722783},
publisher = {Association for Computing Machinery},
address = {New York, NY, USA},
url = {https://doi.org/10.1145/3772318.3790922},
doi = {10.1145/3772318.3790922},
booktitle = {Proceedings of the 2026 CHI Conference on Human Factors in Computing Systems},
articleno = {618},
numpages = {24},
location = {
},
series = {CHI '26}
}

@InProceedings{heo2024rotary,
author="Heo, Byeongho
and Park, Song
and Han, Dongyoon
and Yun, Sangdoo",
editor="Leonardis, Ale{\v{s}}
and Ricci, Elisa
and Roth, Stefan
and Russakovsky, Olga
and Sattler, Torsten
and Varol, G{\"u}l",
title="Rotary Position Embedding for Vision Transformer",
booktitle="Computer Vision -- ECCV 2024",
year="2025",
publisher="Springer Nature Switzerland",
address="Cham",
pages="289--305",
isbn="978-3-031-72684-2"
}

@InProceedings{tomas2021goo,
    author    = {Tomas, Henri and Reyes, Marcus and Dionido, Raimarc and Ty, Mark and Mirando, Jonric and Casimiro, Joel and Atienza, Rowel and Guinto, Richard},
    title     = {GOO: A Dataset for Gaze Object Prediction in Retail Environments},
    booktitle = {Proceedings of the IEEE/CVF Conference on Computer Vision and Pattern Recognition (CVPR) Workshops},
    month     = {June},
    year      = {2021},
    pages     = {3125-3133}
}

@InProceedings{hu2023gfie,
    author    = {Hu, Zhengxi and Yang, Yuxue and Zhai, Xiaolin and Yang, Dingye and Zhou, Bohan and Liu, Jingtai},
    title     = {GFIE: A Dataset and Baseline for Gaze-Following From 2D to 3D in Indoor Environments},
    booktitle = {Proceedings of the IEEE/CVF Conference on Computer Vision and Pattern Recognition (CVPR)},
    month     = {June},
    year      = {2023},
    pages     = {8907-8916}
}

@misc{zhang2025gaze_target_benchmark,
      title={Vision-Language Models Mistake Head Orientation for Gaze Direction: Nonverbal Conversation Cues}, 
      author={Zory Zhang and Pinyuan Feng and Bingyang Wang and Tianwei Zhao and Suyang Yu and Qingying Gao and Hokin Deng and Ziqiao Ma and Yijiang Li and Dezhi Luo},
      year={2026},
      eprint={2506.05412},
      archivePrefix={arXiv},
      primaryClass={cs.CV},
      url={https://arxiv.org/abs/2506.05412}, 
}

@InProceedings{chong2020vat,
author = {Chong, Eunji and Wang, Yongxin and Ruiz, Nataniel and Rehg, James M.},
title = {Detecting Attended Visual Targets in Video},
booktitle = {IEEE/CVF Conference on Computer Vision and Pattern Recognition (CVPR)},
month = {June},
year = {2020}
}

@InProceedings{gupta2022modular,
    author    = {Gupta, Anshul and Tafasca, Samy and Odobez, Jean-Marc},
    title     = {A Modular Multimodal Architecture for Gaze Target Prediction: Application to Privacy-Sensitive Settings},
    booktitle = {Proceedings of the IEEE/CVF Conference on Computer Vision and Pattern Recognition (CVPR) Workshops},
    month     = {June},
    year      = {2022},
    pages     = {5041-5050}
}

@InProceedings{miao2023patchlevel,
    author    = {Miao, Qiaomu and Hoai, Minh and Samaras, Dimitris},
    title     = {Patch-Level Gaze Distribution Prediction for Gaze Following},
    booktitle = {Proceedings of the IEEE/CVF Winter Conference on Applications of Computer Vision (WACV)},
    month     = {January},
    year      = {2023},
    pages     = {880-889}
}

@InProceedings{tafasca2024sharingan,
    author    = {Tafasca, Samy and Gupta, Anshul and Odobez, Jean-Marc},
    title     = {Sharingan: A Transformer Architecture for Multi-Person Gaze Following},
    booktitle = {Proceedings of the IEEE/CVF Conference on Computer Vision and Pattern Recognition (CVPR)},
    month     = {June},
    year      = {2024},
    pages     = {2008-2017}
}

@inproceedings{recasens2015gazefollow,
 author = {Recasens, Adria and Khosla, Aditya and Vondrick, Carl and Torralba, Antonio},
 booktitle = {Advances in Neural Information Processing Systems},
 editor = {C. Cortes and N. Lawrence and D. Lee and M. Sugiyama and R. Garnett},
 pages = {},
 publisher = {Curran Associates, Inc.},
 title = {Where are they looking?},
 url = {https://proceedings.neurips.cc/paper_files/paper/2015/file/ec8956637a99787bd197eacd77acce5e-Paper.pdf},
 volume = {28},
 year = {2015}
}

@InProceedings{tafasca2023childplay,
    author    = {Tafasca, Samy and Gupta, Anshul and Odobez, Jean-Marc},
    title     = {ChildPlay: A New Benchmark for Understanding Children's Gaze Behaviour},
    booktitle = {Proceedings of the IEEE/CVF International Conference on Computer Vision (ICCV)},
    month     = {October},
    year      = {2023},
    pages     = {20935-20946}
}

@InProceedings{chong2018connecting,
author = {Chong, Eunji and Ruiz, Nataniel and Wang, Yongxin and Zhang, Yun and Rozga, Agata and Rehg, James M.},
title = {Connecting Gaze, Scene, and Attention: Generalized Attention Estimation via Joint Modeling of Gaze and Scene Saliency},
booktitle = {Proceedings of the European Conference on Computer Vision (ECCV)},
month = {September},
year = {2018}
}

@InProceedings{lian2018believe,
author="Lian, Dongze
and Yu, Zehao
and Gao, Shenghua",
editor="Jawahar, C. V.
and Li, Hongdong
and Mori, Greg
and Schindler, Konrad",
title="Believe It or Not, We Know What You Are Looking At!",
booktitle="Computer Vision -- ACCV 2018",
year="2019",
publisher="Springer International Publishing",
address="Cham",
pages="35--50",
isbn="978-3-030-20893-6"
}

@InProceedings{fang2021dualattn,
    author    = {Fang, Yi and Tang, Jiapeng and Shen, Wang and Shen, Wei and Gu, Xiao and Song, Li and Zhai, Guangtao},
    title     = {Dual Attention Guided Gaze Target Detection in the Wild},
    booktitle = {Proceedings of the IEEE/CVF Conference on Computer Vision and Pattern Recognition (CVPR)},
    month     = {June},
    year      = {2021},
    pages     = {11390-11399}
}

@InProceedings{bao2022escnet,
    author    = {Bao, Jun and Liu, Buyu and Yu, Jun},
    title     = {ESCNet: Gaze Target Detection With the Understanding of 3D Scenes},
    booktitle = {Proceedings of the IEEE/CVF Conference on Computer Vision and Pattern Recognition (CVPR)},
    month     = {June},
    year      = {2022},
    pages     = {14126-14135}
}

@article{jin2022depthaware,
title = {Depth-aware gaze-following via auxiliary networks for robotics},
journal = {Engineering Applications of Artificial Intelligence},
volume = {113},
pages = {104924},
year = {2022},
issn = {0952-1976},
doi = {https://doi.org/10.1016/j.engappai.2022.104924},
url = {https://www.sciencedirect.com/science/article/pii/S0952197622001464},
author = {Tianlei Jin and Qizhi Yu and Shiqiang Zhu and Zheyuan Lin and Jie Ren and Yuanhai Zhou and Wei Song}
}

@article{song2024vitgaze,
  author  = {Song, Yuehao and Wang, Xinggang and Yao, Jingfeng and Liu, Wenyu and Zhang, Jinglin and Xu, Xiangmin},
  title   = {{ViTGaze}: Gaze Following with Interaction Features in Vision Transformers},
  journal = {Visual Intelligence},
  year    = {2024},
  volume  = {2},
  number  = {1},
  pages   = {31},
  doi     = {10.1007/s44267-024-00064-9},
  url     = {https://doi.org/10.1007/s44267-024-00064-9},
  issn    = {2731-9008}
}

@article{simeoni2025dinov3,
  title={Dinov3},
  author={Sim{\'e}oni, Oriane and Vo, Huy V and Seitzer, Maximilian and Baldassarre, Federico and Oquab, Maxime and Jose, Cijo and Khalidov, Vasil and Szafraniec, Marc and Yi, Seungeun and Ramamonjisoa, Micha{\"e}l and others},
  journal={arXiv preprint arXiv:2508.10104},
  year={2025}
}

@article{oquab2024dinov2,
title={{DINO}v2: Learning Robust Visual Features without Supervision},
author={Maxime Oquab and Timoth{\'e}e Darcet and Th{\'e}o Moutakanni and Huy V. Vo and Marc Szafraniec and Vasil Khalidov and Pierre Fernandez and Daniel HAZIZA and Francisco Massa and Alaaeldin El-Nouby and Mido Assran and Nicolas Ballas and Wojciech Galuba and Russell Howes and Po-Yao Huang and Shang-Wen Li and Ishan Misra and Michael Rabbat and Vasu Sharma and Gabriel Synnaeve and Hu Xu and Herve Jegou and Julien Mairal and Patrick Labatut and Armand Joulin and Piotr Bojanowski},
journal={Transactions on Machine Learning Research},
issn={2835-8856},
year={2024},
url={https://openreview.net/forum?id=a68SUt6zFt},
note={Featured Certification}
}

@InProceedings{cao2026tipsv2,
    author    = {Cao, Bingyi and Chen, Koert and Maninis, Kevis-Kokitsi and Chen, Kaifeng and Karpur, Arjun and Xia, Ye and Dua, Sahil and Dabral, Tanmaya and Han, Guangxing and Han, Bohyung and Ainslie, Joshua and Bewley, Alex and Jacob, Mithun and Wagner, Ren\'e and Ramos, Washington and Choromanski, Krzysztof and Seyedhosseini, Mojtaba and Zhou, Howard and Araujo, Andre},
    title     = {TIPSv2: Advancing Vision-Language Pretraining with Enhanced Patch-Text Alignment},
    booktitle = {Proceedings of the IEEE/CVF Conference on Computer Vision and Pattern Recognition (CVPR)},
    month     = {June},
    year      = {2026},
    pages     = {29325-29335}
}

@InProceedings{radford2021clip,
  title = 	 {Learning Transferable Visual Models From Natural Language Supervision},
  author =       {Radford, Alec and Kim, Jong Wook and Hallacy, Chris and Ramesh, Aditya and Goh, Gabriel and Agarwal, Sandhini and Sastry, Girish and Askell, Amanda and Mishkin, Pamela and Clark, Jack and Krueger, Gretchen and Sutskever, Ilya},
  booktitle = 	 {Proceedings of the 38th International Conference on Machine Learning},
  pages = 	 {8748--8763},
  year = 	 {2021},
  editor = 	 {Meila, Marina and Zhang, Tong},
  volume = 	 {139},
  series = 	 {Proceedings of Machine Learning Research},
  month = 	 {18--24 Jul},
  publisher =    {PMLR},
  url = 	 {https://proceedings.mlr.press/v139/radford21a.html}
}

@inproceedings{chen2023vitadapter,
title={Vision Transformer Adapter for Dense Predictions},
author={Zhe Chen and Yuchen Duan and Wenhai Wang and Junjun He and Tong Lu and Jifeng Dai and Yu Qiao},
booktitle={The Eleventh International Conference on Learning Representations },
year={2023},
url={https://openreview.net/forum?id=plKu2GByCNW}
}

@InProceedings{tonini2023objectaware,
    author    = {Tonini, Francesco and Dall'Asen, Nicola and Beyan, Cigdem and Ricci, Elisa},
    title     = {Object-aware Gaze Target Detection},
    booktitle = {Proceedings of the IEEE/CVF International Conference on Computer Vision (ICCV)},
    month     = {October},
    year      = {2023},
    pages     = {21860-21869}
}

@InProceedings{tu2022end2end,
    author    = {Tu, Danyang and Min, Xiongkuo and Duan, Huiyu and Guo, Guodong and Zhai, Guangtao and Shen, Wei},
    title     = {End-to-End Human-Gaze-Target Detection With Transformers},
    booktitle = {Proceedings of the IEEE/CVF Conference on Computer Vision and Pattern Recognition (CVPR)},
    month     = {June},
    year      = {2022},
    pages     = {2202-2210}
}

@misc{tu2023joint,
      title={Joint Gaze-Location and Gaze-Object Detection}, 
      author={Danyang Tu and Wei Shen and Wei Sun and Xiongkuo Min and Guangtao Zhai},
      year={2023},
      eprint={2308.13857},
      archivePrefix={arXiv},
      primaryClass={cs.CV},
      url={https://arxiv.org/abs/2308.13857}, 
}

@InProceedings{frank2025kddataset,
    author    = {Frank, Logan and Davis, Jim},
    title     = {What Makes a Good Dataset for Knowledge Distillation?},
    booktitle = {Proceedings of the IEEE/CVF Conference on Computer Vision and Pattern Recognition (CVPR)},
    month     = {June},
    year      = {2025},
    pages     = {23755-23764}
}

@InProceedings{andriluka2014mpii,
author = {Andriluka, Mykhaylo and Pishchulin, Leonid and Gehler, Peter and Schiele, Bernt},
title = {2D Human Pose Estimation: New Benchmark and State of the Art Analysis},
booktitle = {Proceedings of the IEEE Conference on Computer Vision and Pattern Recognition (CVPR)},
month = {June},
year = {2014}
}

@article{krasin2017openimages,
  title={OpenImages: A public dataset for large-scale multi-label and multi-class image classification.},
  author={Krasin, Ivan and Duerig, Tom and Alldrin, Neil and Ferrari, Vittorio and Abu-El-Haija, Sami and Kuznetsova, Alina and Rom, Hassan and Uijlings, Jasper and Popov, Stefan and Kamali, Shahab and Malloci, Matteo and Pont-Tuset, Jordi and Veit, Andreas and Belongie, Serge and Gomes, Victor and Gupta, Abhinav and Sun, Chen and Chechik, Gal and Cai, David and Feng, Zheyun and Narayanan, Dhyanesh and Murphy, Kevin},
  journal={Dataset available from https://storage.googleapis.com/openimages/web/index.html},
  year={2017}
}

@InProceedings{lin2014coco,
author="Lin, Tsung-Yi
and Maire, Michael
and Belongie, Serge
and Hays, James
and Perona, Pietro
and Ramanan, Deva
and Doll{\'a}r, Piotr
and Zitnick, C. Lawrence",
editor="Fleet, David
and Pajdla, Tomas
and Schiele, Bernt
and Tuytelaars, Tinne",
title="Microsoft COCO: Common Objects in Context",
booktitle="Computer Vision -- ECCV 2014",
year="2014",
publisher="Springer International Publishing",
address="Cham",
pages="740--755",
isbn="978-3-319-10602-1"
}

@InProceedings{cao2026anygaze,
    author    = {Cao, Xu and Yang, Houze and Gunda, Vipin and Zhou, Zhongyi and Xu, Tianyu and Kowdle, Adarsh and Kim, Inki and Rehg, James M.},
    title     = {Gaze Target Estimation Anywhere with Concepts},
    booktitle = {Proceedings of the IEEE/CVF Conference on Computer Vision and Pattern Recognition (CVPR)},
    month     = {June},
    year      = {2026},
    pages     = {31304-31315}
}

@article{admoni2017social,
  title={Social eye gaze in human-robot interaction: a review},
  author={Admoni, Henny and Scassellati, Brian},
  journal={Journal of Human-Robot Interaction},
  volume={6},
  number={1},
  pages={25--63},
  year={2017},
  publisher={Journal of Human-Robot Interaction Steering Committee}
}

@inproceedings{10.1145/3772318.3790345,
author = {Zhang, Chengwen and Yu, Chun and Zhuang, Borong and Jin, Haopeng and Wan, Qingyang and Li, Zhuojun and He, Zhe and Ye, Zhoutong and Mei, Yu and Liu, Chang and Shi, Weinan and Shi, Yuanchun},
title = {HiSync: Spatio-Temporally Aligning Hand Motion from Wearable IMU and On-Robot Camera for Command Source Identification in Long-Range HRI},
year = {2026},
isbn = {9798400722783},
publisher = {Association for Computing Machinery},
address = {New York, NY, USA},
url = {https://doi.org/10.1145/3772318.3790345},
doi = {10.1145/3772318.3790345},
booktitle = {Proceedings of the 2026 CHI Conference on Human Factors in Computing Systems},
articleno = {1036},
numpages = {21},
location = {
},
series = {CHI '26}
}

@misc{palider2025gazeestimationhumanrobotinteraction,
      title={Gaze Estimation for Human-Robot Interaction: Analysis Using the NICO Platform}, 
      author={Matej Palider and Omar Eldardeer and Viktor Kocur},
      year={2025},
      eprint={2509.24001},
      archivePrefix={arXiv},
      primaryClass={cs.CV},
      url={https://arxiv.org/abs/2509.24001}, 
}

@misc{liu2026aamultiviewmultimodaldataset,
      title={AA: A Multi-view Multimodal Dataset for Screen-based Gaze Estimation}, 
      author={Chang Liu and Jiaqi Liu and Zhoutong Ye and Xinjie Shen and Chun Yu and Yuanchun Shi},
      year={2026},
      eprint={2606.31211},
      archivePrefix={arXiv},
      primaryClass={cs.CV},
      url={https://arxiv.org/abs/2606.31211}, 
}
\bibliographystyle{iclr2026_conference}

\appendix
\section{Detailed Training Procedures}
We provide the technical details of how we trained the teacher and student models. All training stages use the same data augmentation pipeline as Gaze-LLE's (random cropping, horizontal flipping, and bounding box jittering). To speed up training, we train all models in BF16 using PyTorch's Automatic Mixed Precision (AMP) training. The models were trained on a single Nvidia H100 GPU. Finetuning / distilling ViT-L and ViT-H+ models requires much more VRAM than the 80GB available, so we split the batches into smaller ones (30 for ViT-L, 12 for ViT-H+), and accumulate the gradient to achieve an effective batch size of 60.

\subsection{Training the Teacher Model}
\label{sec:appendix-training}
As laid out in Section \ref{sec:method-arch}, we train the teacher model in 2 stages. We first train the decoder only and then finetune the full model (backbone included).

\textbf{Stage 1: Decoder Training.} The decoder is trained using AdamW with a weight decay of $0.05$ and a 100-iteration linear LR warmup from $\eta_{0}=10^{-4}$. The rest of the recipe is the same as Gaze-LLE's: we train the model for $15$ epochs with a batch size of $60$ and an initial learning rate of $\eta=10^{-3}$ after warmup, and use a cosine LR scheduler with $\eta_{min}=10^{-7}$. 

\textbf{Stage 2: Finetuning.} In this stage, the full model is trained with the AdamW optimizer for 5 epochs with a weight decay of $0.01$ and a cosine LR schedule with $\eta_{min}=10^{-7}$. We observe that this process is susceptible to overfitting and training instability. Therefore, we use a long 500-iteration linear warmup from $\eta_{0}=10^{-7}$, a low post-warmup initial learning rate of $\eta=10^{-5}$, and a more aggressive $p=0.5$ channel dropout for backbone features.

\subsection{Distillation}
\label{sec:appendix-distillation}
Similar to the teacher model, we train the student model in two stages as well.

\textbf{Stage 1: Feature Distillation.} We train the student model with AdamW, using a weight decay of $0.01$, an initial LR of $\eta=2\times10^{-4}$ and a cosine LR schedule with $\eta_{min}=10^{-7}$. The distillation runs for 20 epochs (14 more than what is needed to make a good vodka;). 

\textbf{Stage 2: Finetuning.} Since the student only learned feature representation and not downstream prediction during distillation, we adopt a supervised finetuning regime on the labeled train set to fully adapt the student to the downstream task. Specifically, we initialize the student heatmap and in/out heads with teacher weights and finetune the model for 3 epochs. We use a post-warmup initial LR of $\eta=2\times10^{-5}$. The rest of the finetuning regime is identical to the one in Appendix \ref{sec:appendix-training}.

\section{Impact of VAT and ChildPlay Frame Sampling}
\label{sec:appendix-train-set-ablation}

\begin{table}[htbp]
\scriptsize
\centering
\setlength{\tabcolsep}{4pt}
\begin{tabular}{lccccccccccc}
\toprule
\multirow{2}{*}{Experiment}                                                                         & \multirow{2}{*}{\begin{tabular}[c]{@{}c@{}}VAT\\ Frames\end{tabular}} & \multirow{2}{*}{\begin{tabular}[c]{@{}c@{}}ChildPlay\\ Frames\end{tabular}} & \multicolumn{3}{c}{GazeFollow}                      & \multicolumn{3}{c}{VideoAttentionTarget}            & \multicolumn{3}{c}{ChildPlay}                       \\
                                                                                                    &                                                                       &                                                                             & AUC             & Avg L2          & Min L2          & AUC             & L2              & AP{\tiny in/out}              & AUC             & L2              & AP{\tiny in/out}              \\ \midrule
GazeFollow-only                                                                                     & Not Used                                                              & Not Used                                                                    & 0.9595          & 0.0958          & 0.0391          & 0.9488          & {\ul 0.0870}    & N/A             & 0.9572          & 0.0963          & N/A             \\ \midrule
\begin{tabular}[c]{@{}l@{}}Labeled Train Set\\ (Main Experiment)\end{tabular}                       & 1/3                                                                   & 1/6                                                                         & \textbf{0.9599} & {\ul 0.0939}    & {\ul 0.0380}    & 0.9557          & 0.0886          & \textbf{0.9421} & \textbf{0.9652} & \textbf{0.0883} & \textbf{0.9961} \\ \midrule
\multirow{3}{*}{\begin{tabular}[c]{@{}l@{}}Alternative VAT\\ Sampling Intervals\end{tabular}}       & All                                                                   & 1/6                                                                         & 0.9594          & \textbf{0.0928} & \textbf{0.0373} & 0.9552          & 0.0889          & {\ul 0.9360}    & 0.9627          & 0.0942          & 0.9941          \\
                                                                                                    & 1/6                                                                   & 1/6                                                                         & {\ul 0.9598}    & 0.0949          & 0.0391          & 0.9542          & 0.0900          & 0.9345          & {\ul 0.9647}    & {\ul 0.0902}    & 0.9950          \\
                                                                                                    & 1/12                                                                  & 1/6                                                                         & 0.9595          & 0.0956          & 0.0384          & 0.9544          & \textbf{0.0859} & 0.9246          & 0.9634          & 0.0924          & 0.9947          \\ \midrule
\multirow{3}{*}{\begin{tabular}[c]{@{}l@{}}Alternative ChildPlay\\ Sampling Intervals\end{tabular}} & 1/3                                                                   & All                                                                         & 0.9595          & 0.0946          & 0.0385          & \textbf{0.9577} & 0.0898          & 0.9305          & 0.9628          & 0.0954          & 0.9926          \\
                                                                                                    & 1/3                                                                   & 1/3                                                                         & 0.9596          & 0.0965          & 0.0402          & 0.9547          & 0.0906          & 0.9342          & 0.9627          & 0.0960          & 0.9929          \\
                                                                                                    & 1/3                                                                   & 1/12                                                                        & {\ul 0.9598}    & 0.0955          & 0.0390          & {\ul 0.9570}    & 0.0882          & 0.9332          & 0.9634          & 0.0910          & {\ul 0.9952}    \\ \bottomrule
\end{tabular}
\caption{The impact of different train set composition. Best performance is \textbf{bolded} while the runner-up is {\ul underlined}. The composition we chose for our labeled train set (all GazeFollow samples + 1 out of every 3 VAT frames + 1 out of every 6 ChildPlay frames) yields the best overall result.}
\label{tab:appendix-recipe-ablation}
\end{table}

We conducted an ablation study on the composition of the labeled train set to validate our decision to sample VAT and ChildPlay frames at a reduced rate (Section \ref{sec:method-training}). We train \Name{} ViT-B (w/o finetuning) on a variety of data mixtures and report the results in Table \ref{tab:appendix-recipe-ablation}. Sampling 1 out of 3 VAT frames and 1 out of 6 ChildPlay frames yields the best results. As a result, we choose that as the default composition of our labeled train set.

\section{Alternative Backbones}
\label{sec:appendix-backbone-ablation}
We explored using different backbones for \Name{}. Specifically, we compare 4 ViT-L backbones—DINOv2, CLIP, TIPSv2, and DINOv3. We train \Name{} ViT-L (w/o finetuning) using each of these backbones and report the results in Table \ref{tab:appendix-backbone-ablation}. \Name{} is capable of SOTA performance regardless of which backbone is used, although DINOv3 performed the best. This portability across backbones makes it likely that \Name{} can continue to benefit from progress in general-purpose vision backbones.

\begin{table}[htbp]
\scriptsize
\centering
\begin{tabular}{lccccccccc}
\toprule
\multicolumn{1}{c}{\multirow{2}{*}{Backbone}} & \multicolumn{3}{c}{GazeFollow} & \multicolumn{3}{c}{VideoAttentionTarget} & \multicolumn{3}{c}{ChildPlay}  \\
\multicolumn{1}{c}{}                          & AUC      & Avg L2   & Min L2   & AUC        & L2        & AP{\tiny in/out}& AUC    & L2     & AP{\tiny in/out} \\ \midrule
DINOv2~\citep{oquab2024dinov2}                                        & 0.9606   & 0.0895   & 0.0361   & 0.9586     & 0.0856    & 0.9260          & 0.9673 & 0.0842 & 0.9956       \\
CLIP~\citep{radford2021clip}                                          & 0.9604   & 0.0910   & 0.0373   & 0.9625     & 0.0837    & 0.9298          & 0.9667 & 0.0859 & \textbf{0.9966}       \\
TIPSv2~\citep{cao2026tipsv2}                                        & 0.9606   & 0.0925   & 0.0361   & 0.9529     & 0.0920    & 0.9386          & 0.9648 & 0.0850 & 0.9958       \\ 
DINOv3~\citep{simeoni2025dinov3}                                        & \textbf{0.9623}   & \textbf{0.0881}   & \textbf{0.0338}   & \textbf{0.9643}     & \textbf{0.0804}    & \textbf{0.9449}          & \textbf{0.9678} & \textbf{0.0798} & 0.9950       \\ \midrule
DINOv3 + CLIP                                 & 0.9620   & 0.0878   & 0.0339   & 0.9652     & 0.0801    & 0.9368          & 0.9690 & 0.0796 & 0.9957       \\ \bottomrule
\end{tabular}
\caption{Performance of \Name{} with different ViT-L backbones. All models were trained on the labeled train set without finetuning.}
\label{tab:appendix-backbone-ablation}
\end{table}

Considering that gaze is inherently social and is extensively described in text, we hypothesize that models trained on vision language (VL) tasks like CLIP and TIPSv2 could provide additional features that complement DINOv3. We further explore concatenating DINOv3 and CLIP feature maps along the \textit{channel} dimension as the input to a \Name{} decoder. Specifically, we use $512\times512$ scene input for DINOv3 ViT-L ($16\times16$ patches) and $448\times448$ scene input for CLIP ViT-L ($14\times14$ patches). This generates two $32\times32\times768$ feature maps that can be directly concatenated to form one $32\times32\times1536$ feature map. Similarly, we resize the head crop to $256\times256$ for DINOv3 and $224\times224$ for CLIP, and concatenate the head feature maps along the channel dimension. 

We report the performance of this hybrid backbone in Table \ref{tab:appendix-backbone-ablation}. The hybrid backbone slightly improves performance but has $2\times$ the computational cost, so we did not use the hybrid backbone in our main experiments. As VL backbones continue to evolve, we could expect hybrid backbones to bring larger improvements to gaze estimation that justify the extra FLOPs in the near future. We leave such explorations to future work.

\section{2D RoPE for Cross Attention with Unified Coordinates}
\label{sec:appendix-rope-formulation}

We provide the formal definition of 2D RoPE with unified global-local coordinates used in SIM's cross attention where we model the interaction between scene branch and head branch features. Let a patch token at grid position \(p_n=(y_n,x_n)\). For a head with
rotary dimension \(d_r\), assume \(d_r\) is divisible by \(4\), and write
the rotary part of each query/key as
\[
q_n = (q^y_n, q^x_n, q^\perp_n), \qquad
k_n = (k^y_n, k^x_n, k^\perp_n),
\]
where \(q^y_n,q^x_n,k^y_n,k^x_n\in\mathbb{C}^{d_r/4}\), using adjacent
real channels as complex numbers. Define
\[
\theta_i = \theta^{-i/(d_r/4)}, \qquad i=0,\ldots,d_r/4-1.
\]
Following \citet{heo2024rotary}, we use \(\theta=100\) by default. See Appendix \ref{sec:appendix-rope-theta} for results with alternative $\theta$ values.

For self-attention, axial 2D RoPE applies
\[
\widetilde q^y_{n,i}=q^y_{n,i} e^{\mathrm{i}\theta_i y_n}, \qquad
\widetilde q^x_{n,i}=q^x_{n,i} e^{\mathrm{i}\theta_i x_n},
\]
\[
\widetilde k^y_{n,i}=k^y_{n,i} e^{\mathrm{i}\theta_i y_n}, \qquad
\widetilde k^x_{n,i}=k^x_{n,i} e^{\mathrm{i}\theta_i x_n}.
\]
Then attention is computed as
\[
\operatorname{Attn}(Q,K,V)
=
\operatorname{softmax}
\left(
\frac{\widetilde Q \widetilde K^\top}{\sqrt{d_h}}
\right)V.
\]

For cross-attention, let the global scene grid be \(H_g\times W_g\), and
let the local head crop grid be \(H_\ell\times W_\ell\). Given a head bounding
box
\[
x_{\mathrm{bbox}}=(y_0,x_0,y_1,x_1)\in\mathbb{R}^4
\]
in global-grid coordinates, assign local patch \((a,b)\) the continuous
global coordinate
\[
\phi_\ell(a,b)
=
\left(
y_0 + \frac{a+\frac12}{H_\ell}(y_1-y_0)-\frac12,\;
x_0 + \frac{b+\frac12}{W_\ell}(x_1-x_0)-\frac12
\right).
\]
Global-scene patches use their native coordinates
\[
\phi_g(a,b)=(a,b).
\]

For cross-attention from stream \(A\) to stream \(B\), rotate queries using
\(\phi_A\), rotate keys using \(\phi_B\), and leave values unrotated:
\[
\operatorname{CrossAttn}(X_A,X_B)
=
\operatorname{softmax}
\left(
\frac{\widetilde Q_A \widetilde K_B^\top}{\sqrt{d_h}}
\right)V_B.
\]
Thus, for local-to-global attention, we use \(\phi_A=\phi_\ell\),
\(\phi_B=\phi_g\). For global-to-local attention, we use
\(\phi_A=\phi_g\), \(\phi_B=\phi_\ell\). This puts both branches in a unified coordinate system and implicitly encodes head location, allowing us to remove Gaze-LLE's learnable head prompt.

\section{Alternative SIM Layer Designs}
\label{sec:appendix-layer-ablation}

\begin{figure}[h!]
  \centering
  \includegraphics[width=0.60\linewidth]{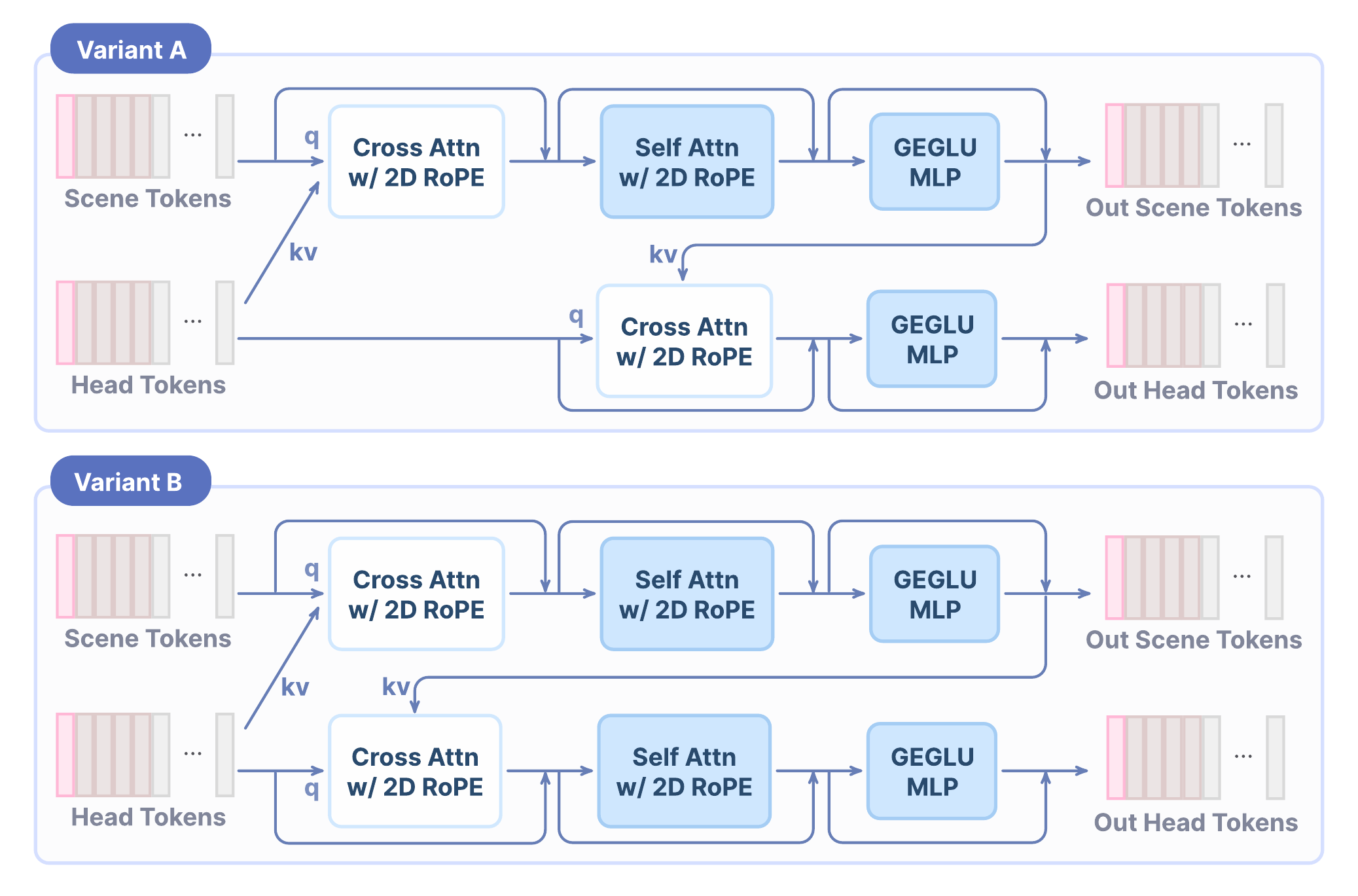}
  \caption{Two other SIM variants we explored. Both performed worse than our eventual architecture (see Figure \ref{fig:architecture}).}
  \label{fig:appendix-alternative-sim}
\end{figure}

We explored two other architectures (Figure \ref{fig:appendix-alternative-sim}) for modeling the interaction between the scene and head branches. We train a \Name{} ViT-B (w/o finetuning) with each of the designs and report the results in Table \ref{tab:appendix-layer-ablation}. Coincidentally, Variant A is architecturally similar to ViTAdapter~\citep{chen2023vitadapter}, which uses cross attention to model the interaction between a feature pyramid derived from convolutions and ViT features for dense prediction tasks. Our eventual SIM architecture, which includes an additional self attention for the head branch, leads to the best overall results.

\begin{table}[h!]
\centering
\scriptsize
\begin{tabular}{lccccccccc}
\toprule
\multirow{2}{*}{Experiment} & \multicolumn{3}{c}{GazeFollow}                      & \multicolumn{3}{c}{VideoAttentionTarget}            & \multicolumn{3}{c}{ChildPlay}                       \\
                            & AUC             & Avg L2          & Min L2          & AUC             & L2              & AP{\tiny in/out}              & AUC             & L2              & AP{\tiny in/out}              \\ \midrule
Variant A                   & 0.9595          & 0.0971          & 0.0402          & 0.9554          & 0.0883 & 0.9350          & 0.9617          & 0.0942          & 0.9952          \\
Variant B                   & 0.9593          & 0.0962          & 0.0396          & \textbf{0.9562} & \textbf{0.0875}          & 0.9277          & 0.9623          & 0.0981          & 0.9925         \\ 
SIM                         & \textbf{0.9599} & \textbf{0.0939} & \textbf{0.0380} & 0.9557          & 0.0886          & \textbf{0.9421} & \textbf{0.9652} & \textbf{0.0883} & \textbf{0.9961} \\ \bottomrule
\end{tabular}
\caption{Comparison between SIM and two variants.}
\label{tab:appendix-layer-ablation}
\end{table}

\section{More Ablation Studies of the \Name{} Architecture}
\label{sec:appendix-more-decoder-ablation}
We conduct additional ablation studies on decoder hyperparameters and explore different model depths, token dimensions, FFN MLP ratios, FFN types, backbone feature dropout $p$, 2D RoPE $\theta$, and learning rates. We run the experiments with \Name{} ViT-B (w/o finetuning).

\subsection{Model Depth and Width}
\label{sec:appendix-depth-width-ablation}
We report the performance of \Name{} models with varying depth, token dimension, and FFN MLP ratio in Tables \ref{tab:appendix-depth-ablation}, \ref{tab:appendix-width-ablation} and \ref{tab:appendix-mlp-ratio-ablation}. The final combination of these hyperparameters we used in our paper represents a balance between different performance metrics and consistency with prior work.

\begin{table}[h!]
\scriptsize
\centering
\begin{tabular}{lccccccccc}
\toprule
\multirow{2}{*}{\begin{tabular}[c]{@{}l@{}}\# SIM\\ Layers\end{tabular}} & \multicolumn{3}{c}{GazeFollow}                                                    & \multicolumn{3}{c}{VideoAttentionTarget}                                  & \multicolumn{3}{c}{ChildPlay}                                             \\
                                                                        & \multicolumn{1}{c}{AUC} & \multicolumn{1}{c}{Avg L2} & \multicolumn{1}{c}{Min L2} & \multicolumn{1}{c}{AUC} & \multicolumn{1}{c}{L2} & \multicolumn{1}{c}{AP{\tiny in/out}} & \multicolumn{1}{c}{AUC} & \multicolumn{1}{c}{L2} & \multicolumn{1}{c}{AP{\tiny in/out}} \\ \midrule
3 Layers                                                                & {\ul 0.9598}            & 0.0956                     & 0.0389                     & 0.9543                  & 0.0882                 & 0.9285                 & 0.9644                  & \textbf{0.0875}        & 0.9938                 \\
4 Layers                                                                & 0.9597                  & 0.0944                     & {\ul 0.0381}               & 0.9557                  & \textbf{0.0848}        & 0.9256                 & 0.9639                  & 0.0927                 & 0.9932                 \\
5 Layers                                                                & \textbf{0.9599}         & {\ul 0.0939}               & \textbf{0.0380}            & 0.9557                  & 0.0886                 & \textbf{0.9421}        & \textbf{0.9652}         & {\ul 0.0883}           & \textbf{0.9961}        \\
6 Layers                                                                & \textbf{0.9599}         & 0.0957                     & 0.0386                     & \textbf{0.9564}         & 0.0883                 & 0.9259                 & 0.9622                  & 0.0919                 & 0.9949                 \\
7 Layers                                                                & \textbf{0.9599}         & \textbf{0.0935}            & 0.0383                     & {\ul 0.9560}            & {\ul 0.0855}           & {\ul 0.9390}           & 0.9627                  & 0.0941                 & 0.9930                 \\ \bottomrule
\end{tabular}
\caption{\Name{} performance with different numbers of SIM layers. 5 layers leads to the best overall performance, and is the default in this paper.}
\label{tab:appendix-depth-ablation}
\end{table}

\begin{table}[h!]
\scriptsize
\centering
\begin{tabular}{cccccccccc}
\toprule
\multirow{2}{*}{\begin{tabular}[c]{@{}c@{}}Token\\ Dim\end{tabular}} & \multicolumn{3}{c}{GazeFollow}                      & \multicolumn{3}{c}{VideoAttentionTarget}            & \multicolumn{3}{c}{ChildPlay}                       \\
                                                                           & AUC             & Avg L2          & Min L2          & AUC             & L2              & AP{\tiny in/out}              & AUC             & L2              & AP{\tiny in/out}              \\ \midrule
128                                                                        & \textbf{0.9601} & \textbf{0.0933} & {\ul 0.0384}    & \textbf{0.9587} & \textbf{0.0861} & {\ul 0.9408}    & 0.9649          & \textbf{0.0882} & {\ul 0.9950}    \\
192                                                                        & 0.9595          & 0.0952          & 0.0392          & {\ul 0.9580}    & \textbf{0.0861} & 0.9374          & 0.9642          & 0.0888          & 0.9945          \\
256                                                                        & {\ul 0.9599}    & {\ul 0.0939}    & \textbf{0.0380} & 0.9557          & 0.0886          & \textbf{0.9421} & \textbf{0.9652} & {\ul 0.0883}    & \textbf{0.9961} \\
384                                                                        & 0.9590          & 0.1003          & 0.0421          & 0.9490          & 0.0987          & 0.9267          & 0.9596          & 0.0965          & 0.9944          \\ \bottomrule
\end{tabular}
\caption{\Name{} performance with different token dimensions. While there is no significant difference between 128 and 256, we decided to use 256 to stay consistent with \citet{ryan2025gaze-lle}.}
\label{tab:appendix-width-ablation}
\end{table}

\begin{table}[h!]
\scriptsize
\centering
\begin{tabular}{lccccccccc}
\toprule
\multirow{2}{*}{MLP Ratio} & \multicolumn{3}{c}{GazeFollow}                      & \multicolumn{3}{c}{VideoAttentionTarget}            & \multicolumn{3}{c}{ChildPlay}                       \\
                           & AUC             & Avg L2          & Min L2          & AUC             & L2              & AP              & AUC             & L2              & AP              \\ \midrule
2                          & \textbf{0.9601} & 0.0950          & 0.0389          & {\ul 0.9559}    & {\ul 0.0869}    & 0.9235          & {\ul 0.9644}    & {\ul 0.0889}    & {\ul 0.9956}    \\
4                          & 0.9599          & \textbf{0.0939} & \textbf{0.0380} & 0.9557          & 0.0886          & \textbf{0.9421} & \textbf{0.9652} & \textbf{0.0883} & \textbf{0.9961} \\
6                          & 0.9598          & {\ul 0.0943}    & {\ul 0.0381}    & \textbf{0.9588} & \textbf{0.0806} & 0.9238          & 0.9630          & 0.0928          & 0.9937          \\
8                          & \textbf{0.9601} & 0.0954          & 0.0390          & 0.9554          & 0.0880          & {\ul 0.9374}    & 0.9626          & 0.0934          & 0.9945          \\ \bottomrule
\end{tabular}
\caption{\Name{} performance with different MLP ratios. A MLP ratio of 4 leads to the best overall results.}
\label{tab:appendix-mlp-ratio-ablation}
\end{table}

\subsection{GEGLU vs SwiGLU}
\label{sec:appendix-glu}
We report the performance when we substitute GEGLU MLP with SwiGLU MLP in \Name{} in Table \ref{tab:appendix-glu-ablation}. GEGLU performs better than SwiGLU, so we use GEGLU as the default in \Name{}.

\begin{table}[h!]
\scriptsize
\centering
\begin{tabular}{lccccccccc}
\toprule
\multirow{2}{*}{MLP Layer} & \multicolumn{3}{c}{GazeFollow}                      & \multicolumn{3}{c}{VideoAttentionTarget}            & \multicolumn{3}{c}{ChildPlay}                       \\
                           & AUC             & Avg L2          & Min L2          & AUC             & L2              & AP              & AUC             & L2              & AP              \\ \midrule
SwiGLU                     & 0.9595          & 0.0943          & 0.0386          & \textbf{0.9582} & \textbf{0.0857} & 0.9364          & 0.9633          & 0.0935          & 0.9950          \\
GEGLU                      & \textbf{0.9599} & \textbf{0.0939} & \textbf{0.0380} & 0.9557          & 0.0886          & \textbf{0.9421} & \textbf{0.9652} & \textbf{0.0883} & \textbf{0.9961} \\ \bottomrule
\end{tabular}
\caption{\Name{} with different types of MLP layers. GEGLU has the best overall performance and is used as the default.}
\label{tab:appendix-glu-ablation}
\end{table}

\subsection{DINO Feature Dropout Rates}
\label{sec:appendix-dropout}
We report the impact of different DINO feature dropout probabilities ($p$) in Table \ref{tab:appendix-dino-dropout}. We settled on $p=0.1$ because it led to the best overall results.

\begin{table}[h!]
\scriptsize
\centering
\begin{tabular}{lccccccccc}
\toprule
\multirow{2}{*}{\begin{tabular}[c]{@{}l@{}}DINO Feature\\ Dropout $p$\end{tabular}} & \multicolumn{3}{c}{GazeFollow}                      & \multicolumn{3}{c}{VideoAttentionTarget}            & \multicolumn{3}{c}{ChildPlay}                       \\
                                                                                    & AUC             & Avg L2          & Min L2          & AUC             & L2              & AP              & AUC             & L2              & AP              \\ \midrule
0.0                                                                                 & 0.9580          & 0.0963          & 0.0403          & {\ul 0.9564}    & 0.0889          & 0.9264          & {\ul 0.9644}    & 0.0933          & 0.9943          \\
0.1                                                                                 & 0.9599          & \textbf{0.0939} & \textbf{0.0380} & 0.9557          & 0.0886          & \textbf{0.9421} & \textbf{0.9652} & \textbf{0.0883} & \textbf{0.9961} \\
0.3                                                                                 & {\ul 0.9604}    & {\ul 0.0949}    & {\ul 0.0381}    & \textbf{0.9566} & \textbf{0.0856} & {\ul 0.9350}    & 0.9640          & {\ul 0.0917}    & {\ul 0.9957}    \\
0.5                                                                                 & \textbf{0.9609} & 0.0953          & 0.0382          & 0.9540          & {\ul 0.0873}    & 0.9336          & 0.9621          & 0.0918          & 0.9950          \\ \bottomrule
\end{tabular}
\caption{Impact of DINO feature dropout $p$.}
\label{tab:appendix-dino-dropout}
\end{table}

\subsection{2D RoPE Frequency}
\label{sec:appendix-rope-theta}
$\theta$ controls the frequency of 2D RoPE. Both \citet{heo2024rotary} and DINOv3~\citep{simeoni2025dinov3} use $\theta=100$, which is also the default in our paper. We further explore $\theta=10$ and $\theta=10000$ and report the results in Table \ref{tab:appendix-rope-theta-ablation}.

\begin{table}[h!]
\scriptsize
\centering
\begin{tabular}{lccccccccc}
\toprule
\multirow{2}{*}{RoPE $\theta$} & \multicolumn{3}{c}{GazeFollow}                      & \multicolumn{3}{c}{VideoAttentionTarget}            & \multicolumn{3}{c}{ChildPlay}                       \\
                            & AUC             & Avg L2          & Min L2          & AUC             & L2              & AP{\tiny in/out}              & AUC             & L2              & AP{\tiny in/out}              \\ \midrule
10                          & 0.9594          & 0.0948          & 0.0387          & 0.9552          & \textbf{0.0878} & {\ul 0.9321}    & {\ul 0.9650}    & {\ul 0.0885}    & {\ul 0.9954}    \\
100                         & \textbf{0.9599} & \textbf{0.0939} & {\ul 0.0380}    & {\ul 0.9557}    & 0.0886          & \textbf{0.9421} & \textbf{0.9652} & \textbf{0.0883} & \textbf{0.9961} \\
10000                       & {\ul 0.9595}    & {\ul 0.0940}    & \textbf{0.0379} & \textbf{0.9583} & {\ul 0.0882}    & 0.9302          & 0.9628          & 0.0922          & 0.9950          \\ \bottomrule
\end{tabular}
\caption{Effect of different 2D RoPE $\theta$.}
\label{tab:appendix-rope-theta-ablation}
\end{table}

\subsection{Learning Rate}
\label{sec:appendix-learning-rate}
We explore the effect of different learning rates and present the results in Table \ref{tab:appendix-lr-ablation}. We encountered several loss spikes when training the model with $\eta=3\times10^{-3}$, which could be responsible for the abnormally low performance, especially on AP{\tiny in/out} where the model performed similar to blindly guessing ``in frame'' for all samples.

\begin{table}[h!]
\scriptsize
\centering
\begin{tabular}{lccccccccc}
\toprule
\multirow{2}{*}{LR} & \multicolumn{3}{c}{GazeFollow}                      & \multicolumn{3}{c}{VideoAttentionTarget}            & \multicolumn{3}{c}{ChildPlay}                       \\
                    & AUC             & Avg L2          & Min L2          & AUC             & L2              & AP{\tiny in/out}              & AUC             & L2              & AP{\tiny in/out}              \\ \midrule
3e-3                & 0.9486          & 0.1238          & 0.0602          & 0.9218          & 0.1247          & 0.6459          & 0.9419          & 0.1104          & 0.9444          \\
1e-3                & \textbf{0.9599} & \textbf{0.0939} & \textbf{0.0380} & \textbf{0.9557} & {\ul 0.0886}    & \textbf{0.9421} & \textbf{0.9652} & \textbf{0.0883} & \textbf{0.9961} \\
3e-4                & {\ul 0.9577}    & {\ul 0.0954}    & {\ul 0.0398}    & {\ul 0.9555}    & \textbf{0.0857} & {\ul 0.9329}    & {\ul 0.9637}    & {\ul 0.0934}    & {\ul 0.9945}    \\
1e-4                & 0.9573          & 0.0971          & 0.0409          & 0.9552          & 0.0923          & 0.9179          & 0.9630          & 0.0940          & 0.9937          \\ \bottomrule
\end{tabular}
\caption{Effect of different learning rates.}
\label{tab:appendix-lr-ablation}
\end{table}

\section{Distillation Set Data Quality}
\label{sec:appendix-distillation-set-quality}

As laid out in Section \ref{sec:method-distillation-data}, the distillation set obtains head bounding boxes from a YOLO head detector. This automated approach has been validated in previous work~\citep{ryan2025gaze-lle}. We further confirm that the quality of our unlabeled distillation set is on par with the labeled train set through a simple experiment. Since the labeled train set has $200k$ images, we distill 2 student models using the labeled train set and a $200k$ subset of the distillation set, respectively. Both models are then finetuned on the labeled train set per Section \ref{sec:appendix-distillation}. We report the results in Table \ref{tab:appendix-distillation-set-quality}. Both yielded highly similar results, suggesting that our unlabeled distillation set is both high-quality and scalable.

\begin{table}[h!]
\scriptsize
\centering
\begin{tabular}{lccccccccc}
\toprule
\multirow{2}{*}{\begin{tabular}[c]{@{}l@{}}Distillation\\ Data\end{tabular}} & \multicolumn{3}{c}{GazeFollow} & \multicolumn{3}{c}{VideoAttentionTarget}            & \multicolumn{3}{c}{ChildPlay}                       \\
                                                                             & AUC      & Avg L2   & Min L2   & AUC    & L2     & AP{\tiny in/out} & AUC    & L2     & AP{\tiny in/out} \\ \midrule
Labeled Train Set (200k)                                                     & 0.9644   & 0.0830   & 0.0312   & 0.9677 & 0.0720 & 0.9481                            & 0.9730 & 0.0734 & 0.9947                            \\
200k Unlabeled                                                               & 0.9647   & 0.0839   & 0.0316   & 0.9658 & 0.0718 & 0.9484                            & 0.9710 & 0.0728 & 0.9958                            \\ \bottomrule
\end{tabular}
\caption{Distillation with different data. The difference in performance is negligible.}
\label{tab:appendix-distillation-set-quality}
\end{table}

\section{Profiling \Name{} ViT-S Distill for Practical Deployment}
\label{sec:appendix-sft-distill-profiling}
In Section \ref{sec:method-distillation-objective}, we reported the FLOPs of all \Name{} variants, with the distilled models retaining SOTA performance while being lightweight. To further demonstrate that models like \Name{} ViT-S are suitable for practical use cases involving robots or edge devices~\citep{admoni2017social,10.1145/3772318.3790345,palider2025gazeestimationhumanrobotinteraction}, we profile the model on an Nvidia RTX 4090 GPU and an Intel i9-14900K CPU. GPU inference took on average 16ms per frame, while CPU-only inference took around 190ms. This result is achieved using BF16 (PyTorch AMP), which is in line with the training and evaluation procedures used elsewhere in this paper.

\section{Further Analysis of SFT and Distillation}
\label{sec:appendix-sft-distill-ablation}

First, we ablate our distillation objective and present the results in Table \ref{tab:appendix-sft-distill-ablation}. Adding the cosine term and auxiliary loss affects GazeFollow positively and VAT negatively, while the impact on ChildPlay is minimal. Considering GazeFollow is much more diverse and reflective of downstream applications, we decided to keep the cosine term and auxiliary loss. Meanwhile, using L1 instead of the more common MSE loss brings clear improvements.

\begin{table}[h!]
\scriptsize
\centering
\begin{tabular}{lccccccccc}
\toprule
\multirow{2}{*}{Experiment}    & \multicolumn{3}{c}{GazeFollow} & \multicolumn{3}{c}{VideoAttentionTarget} & \multicolumn{3}{c}{ChildPlay} \\
                               & AUC      & Avg L2   & Min L2   & AUC          & L2          & AP          & AUC      & L2       & AP      \\ \midrule
No Cosine Term (200k)          & {\ul 0.9646}   & 0.0849   & 0.0323   & \textbf{0.9671}       & \textbf{0.0684}      & {\ul 0.9459}      & \textbf{0.9714}   & 0.0731   & \textbf{0.9961}  \\
L1 Term $\rightarrow$ MSE Term (200k) & 0.9603   & 0.0946   & 0.0388   & 0.9609       & 0.0793      & 0.9408      & 0.9674   & 0.0790   & 0.9952  \\
No Auxiliary Loss (200k)       & 0.9643   & {\ul 0.0848}   & {\ul 0.0321}   & {\ul 0.9662}       & {\ul 0.0696}      & 0.9426      & {\ul 0.9713}   & \textbf{0.0724}   & 0.9954  \\ 
$\mathcal{L}_{distill}$ (200k) & \textbf{0.9647}   & \textbf{0.0839}   & \textbf{0.0316}   & 0.9658       & 0.0718      & \textbf{0.9484}      & 0.9710   & {\ul 0.0728}   & {\ul 0.9958}  \\ \bottomrule
\end{tabular}
\caption{Ablation study for the distillation objective. All experiments are done with \Name{} ViT-B.}
\label{tab:appendix-sft-distill-ablation}
\end{table}

We compare the performance of the student model before and after SFT. The before-SFT results were obtained by directly using the teacher model's heatmap and in/out heads on top of the student's final layer features. The results in Table \ref{tab:appendix-post-distill-sft} show that post-distillation SFT has an overall positive impact on performance, especially on ChildPlay.

\begin{table}[h!]
\scriptsize
\centering
\begin{tabular}{lccccccccc}
\toprule
\multirow{2}{*}{Experiment} & \multicolumn{3}{c}{GazeFollow}                      & \multicolumn{3}{c}{VideoAttentionTarget}            & \multicolumn{3}{c}{ChildPlay}                       \\
                            & AUC             & Avg L2          & Min L2          & AUC             & L2              & AP              & AUC             & L2              & AP              \\ \midrule
Distillation Only           & 0.9657          & 0.0817          & 0.0299          & \textbf{0.9702} & 0.0680          & \textbf{0.9493} & 0.9732          & 0.0722          & \textbf{0.9970} \\
Distillation + SFT          & \textbf{0.9660} & \textbf{0.0814} & \textbf{0.0295} & 0.9688          & \textbf{0.0677} & 0.9450          & \textbf{0.9734} & \textbf{0.0697} & 0.9969          \\ \bottomrule
\end{tabular}
\caption{SFT improves overall performance of the distilled ViT-B student model.}
\label{tab:appendix-post-distill-sft}
\end{table}

\section{Evaluating Gemini 3.5 Flash}
\label{sec:appendix-gemini-eval}
In this section, we present the detailed evaluation procedure that led to the Gemini 3.5 Flash results reported in Table \ref{tab:full-results-gf-vat-cp}. We formulate the evaluation task as a VQA task with structured output. We overlay the scene image with the bounding box of the person's head, which acts like a visual prompt. Apart from the image, we also provide the model with the following text prompt:

\begin{tcolorbox}[colback=gray!10, colframe=gray!10, breakable]
\small
Task:

1. Look at the whole image and the specified person's head box.

2. The tested person's head is marked by a red rectangle in the image.

3. Decide whether this person's gaze target is inside the image.

4. If it is inside, output one best gaze target point.

Coordinate system:

- Use a 1000x1000 grid over the image that you see.

- (0, 0) is the top-left corner and (999, 999) is the bottom-right corner.

- The image sent to you is \{sent\_width\}x\{sent\_height\} pixels.

- Head box on the 1000x1000 grid is [xmin, ymin, xmax, ymax] = {bbox}.

Return JSON only, with this exact shape:
\{"inout":"in","x":123,"y":456\}

Rules:

- ``inout'' must be either ``in'' or ``out''.

- If ``inout'' is ``out'', set ``x'' and ``y'' to null.

- If ``inout'' is ``in'', ``x'' and ``y'' must be integers from 0 to 999.

- Do not return explanations, Markdown, or extra keys.

\end{tcolorbox}

The $1000\times1000$ grid coordinate system is in line with Gemini's official documentation. We convert that coordinate into actual coordinates on the image during evaluation. Furthermore, this procedure yields a binary ``in'' or ``out'' classification for in/out prediction. This means that both AP{\tiny in/out} (works well with a probabilistic output) and AUC (designed to evaluate heatmaps) are unsuitable for evaluating Gemini. In contrast, only L2 results are comparable between gaze estimation models and Gemini, so we only report L2 metrics in Section \ref{sec:exp-full-results}. We further report the F1 scores for in/out prediction for both Gemini and \Name{} ViT-H+, which (unlike AP{\tiny in/out}) is directly comparable. Gemini achieved an F1 score of 0.811 on VAT and 0.973 on ChildPlay. \Name{} ViT-H+ achieved an F1 score of 0.897 on VAT and 0.985 on ChildPlay, outperforming Gemini.

\section{More Qualitative Examples}
\label{sec:appendix-qualitative-examples}
\begin{figure}[h!]
  \centering
  \includegraphics[width=0.90\linewidth]{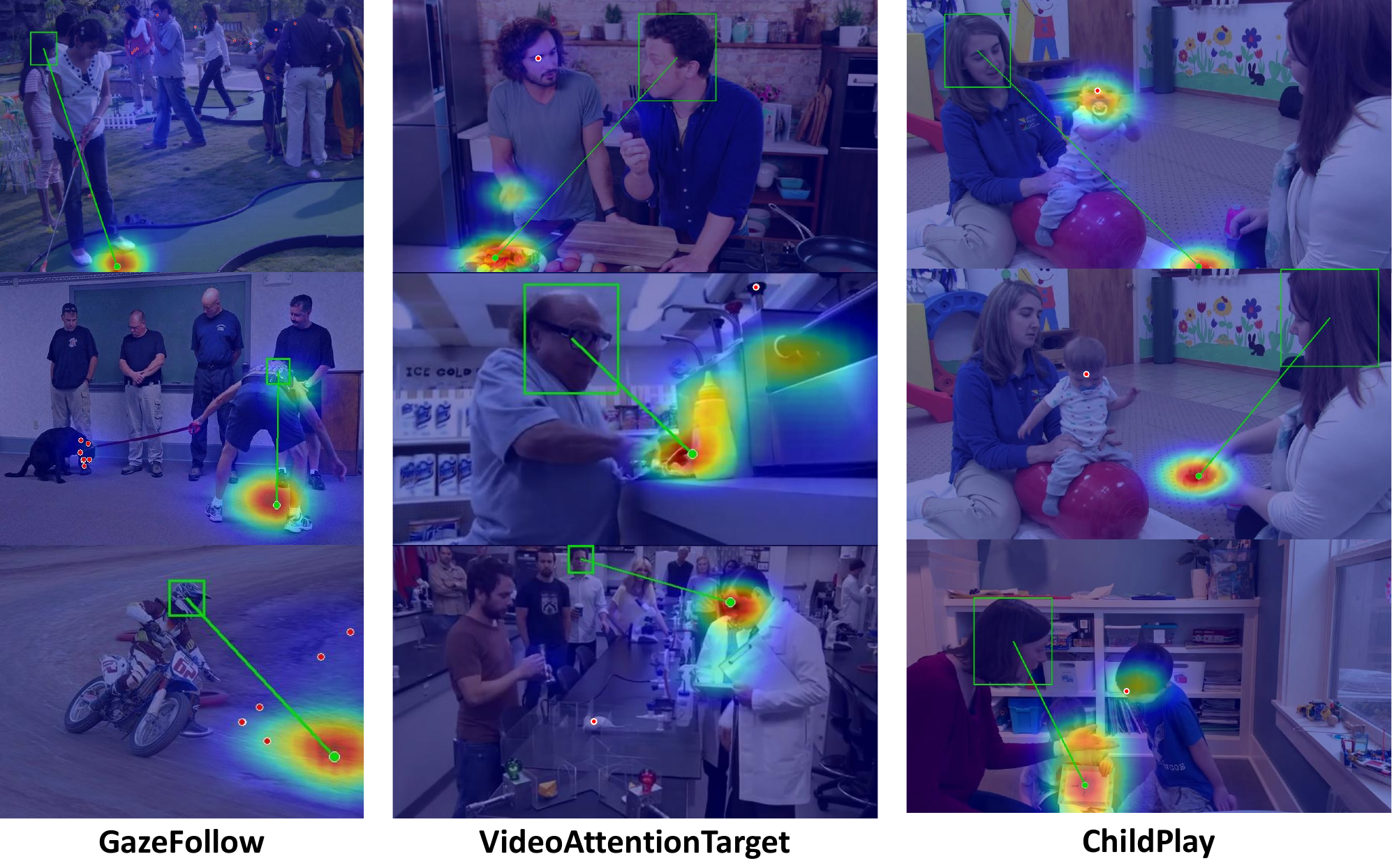}
  \caption{Failure cases where \Name{} performed poorly.}
  \label{fig:appendix-bad-cases}
\end{figure}

We first present cases where \Name{} failed in Figure \ref{fig:appendix-bad-cases}. Many of these cases have occluded faces and eyes, while others depict moments of \textit{gaze shifts}. Blurry images also pose challenges. We expect the model to perform better in at least some of the cases if given enough training data.

\begin{figure}[h!]
  \centering
  \includegraphics[width=0.90\linewidth]{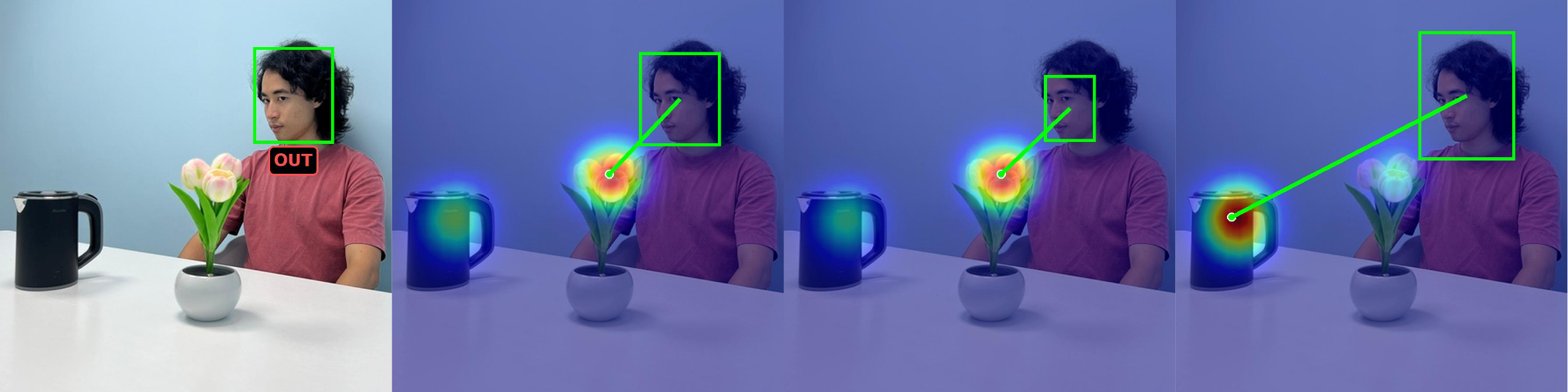}
  \caption{A case from \citet{zhang2025gaze_target_benchmark}'s dataset sensitive to changes in the head bounding box. All 4 bounding boxes shown here are plausible and unambiguous, but lead to different predictions.}
  \label{fig:appendix-bbox-sensitivity}
\end{figure}

In addition, we observe that, in rare cases, \Name{} is sensitive to small variations in the head bounding box. We present an example in Figure \ref{fig:appendix-bbox-sensitivity}. While we find these cases to be very infrequent, future work should focus on mitigating this sensitivity, probably through stronger data augmentation.

\begin{figure}[h!]
  \centering
  \includegraphics[width=0.90\linewidth]{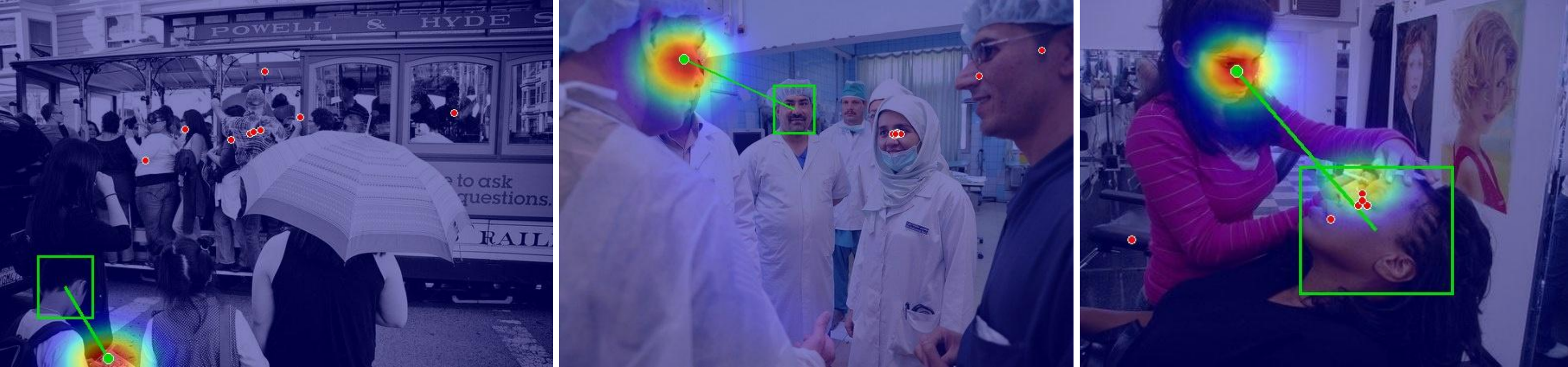}
  \caption{GazeFollow test split samples with incorrect annotations. A common theme is annotators mistakenly labeling the gaze target of an individual unrelated to the head bounding box.}
  \label{fig:appendix-bad-annotations}
\end{figure}

In Figure \ref{fig:appendix-bad-annotations}, we present some cases where the human annotation is unreliable and \Name{}'s prediction makes more sense. This underscores the importance of future datasets providing actual ground truth instead of relying on human annotators.

\section{High Resolution Heatmap Output}
\label{sec:appendix-512px-heatmap}
As discussed in Section \ref{sec:exp-eval-bias}, the discretization errors stemming from the low resolution $64\times64$ heatmap used in standard evaluation protocols are unfair to gaze estimation models when comparing their performance to that of humans. While the fix we introduced in Section \ref{sec:exp-eval-bias} removed a systematic bias towards the top left of the image, the underlying cause, discretization, remains. Here, we further explore the performance impact of training a \Name{} ViT-H+ to output $512\times512$ heatmaps that cause negligible discretization errors.

We replicate the training regime for the \Name{} ViT-H+ teacher model, but this time the model outputs $512\times512$ heatmaps, and we use $512\times512$ ground truth heatmaps as supervision. The increased resolution causes training instabilities, so we use a lower $3\times10^{-4}$ learning rate instead of $10^{-3}$. The SFT learning rate remains the same. We report the results in Table \ref{tab:appendix-512px-heatmap}. Overall, using high resolution $512\times512$ outputs seems like a valid way to reduce systematic error in evaluation. 

\begin{table}[htbp]
\scriptsize
\centering
\begin{tabular}{lccccccccc}
\toprule
\multirow{2}{*}{\begin{tabular}[c]{@{}l@{}}Heatmap\\ Resolution\end{tabular}} & \multicolumn{3}{c}{GazeFollow}                      & \multicolumn{3}{c}{VideoAttentionTarget}            & \multicolumn{3}{c}{ChildPlay}                       \\
                                                                              & AUC             & Avg L2          & Min L2          & AUC             & L2              & AP{\tiny in/out}              & AUC             & L2              & AP{\tiny in/out}              \\ \midrule
64$\times$64 Biased                                                           & 0.9659          & 0.0804          & 0.0288          & 0.9719          & 0.0643          & \textbf{0.9509} & 0.9746          & 0.0687          & 0.9954          \\
64$\times$64 Unbiased                                                         & 0.9659          & \textbf{0.0786} & 0.0277          & 0.9719          & \textbf{0.0624} & \textbf{0.9509} & 0.9746          & \textbf{0.0662} & 0.9954          \\
512$\times$512 Biased                                                         & \textbf{0.9661} & 0.0789          & \textbf{0.0273} & \textbf{0.9720} & 0.0672          & 0.9506          & \textbf{0.9759} & \textbf{0.0662} & \textbf{0.9966} \\ \bottomrule
\end{tabular}
\caption{Performance of \Name{} ViT-H+ models trained with different heatmap resolutions.}
\label{tab:appendix-512px-heatmap}
\end{table}

\section{\Name{} and Detection-based Methods}
\label{sec:appendix-detection-based-methods}

\citet{tu2022end2end} used a different formulation for gaze estimation and proposed HGTTR, a DETR-based joint head and gaze target detection model. Given an RGB image $x_{img} \in \mathbb{R}^{3 \times H_{in} \times W_{in}}$ that contains at least one person, the model predicts a fixed number of $N$ gaze instances. A gaze instance $y$ consists of a predicted head bounding box $y_{bbox}\in[0,1]^4$, the probability that the object is indeed a head $y_{class}\in[0,1]$, an in/out prediction $y_{in/out} \in [0,1]$, and a gaze heatmap $y_{H}\in[0,1]^{H_{out} \times W_{out}}$. Similar formulations are also used by \citet{tu2023joint} and \citet{tonini2023objectaware}. This task definition differs from the one we used (Section \ref{sec:method-problem-def}) in that the head bounding box is not part of the input, and is instead predicted by the model (along with its associated gaze heatmap). 

\subsection{Matching Instances at Train Time}
To calculate loss at train time, HGTTR uses Hungarian algorithm to pair each predicted instance with a ground truth instance. The optimal matching for each ground truth instance $y_i$ is the one that minimizes $\mathcal{L}_{match}(y_i,\hat{y}_{pred})$. In HGTTR, $\mathcal{L}_{match}$ is defined as:
$$
\mathcal{L}_{match}=\lambda_1\mathcal{L}_{bbox}+\lambda_2\mathcal{L}_{class}+\lambda_3\mathcal{L}_{in/out} + \lambda_4\mathcal{L}_{heatmap}
$$
where $\mathcal{L}_{bbox}$ is an IoU head box regression loss, $\mathcal{L}_{class}$ and $\mathcal{L}_{in/out}$ are binary classification losses, and $\mathcal{L}_{heatmap}$ is an L2 heatmap loss. Whatever the input, the model always predicts $N$ instances. All methods in this line of work use $N=20$.

\subsection{Matching Instances at Test Time}
\textbf{The matching process at test time makes fair and direct comparison between detection-based methods and \Name{} impossible.} At test time, these detection-based methods use $\mathcal{L}_{match}$ to select the predicted instance for each ground truth instance, and evaluation metrics (AUC, L2, AP{\tiny in/out}) are calculated accordingly. This process gives these methods a fundamental advantage over \Name{} in that \textbf{the ground truth is used to select the optimal prediction that has the least loss for evaluation}. \citet{ryan2025gaze-lle} observed that, since $N=20$ is substantially larger than the usual number of ground truth instances in GazeFollow and VAT, \citet{tonini2023objectaware}'s model often predicts \textit{several} instances with the same head but different gaze targets. The matching process would then choose the instance with the correct prediction for metric calculation, giving the model an edge.

\subsection{Performance Comparison}

\begin{table}[h!]
\scriptsize
\centering
\begin{tabular}{llcccccc}
\toprule
\multirow{2}{*}{\begin{tabular}[c]{@{}l@{}}Ground Truth\\ Gaze Matching\end{tabular}} & \multirow{2}{*}{Model} & \multicolumn{3}{c}{GazeFollow} & \multicolumn{3}{c}{VideoAttentionTarget} \\
                                                                                      &                        & AUC      & Avg L2   & Min L2   & AUC        & L2        & AP              \\ \midrule
\multirow{3}{*}{\Checkmark}                                                                     & \citet{tu2022end2end}               & 0.917    & 0.133    & 0.069    & 0.893      & 0.137     & 0.821           \\
                                                                                      & \citet{tu2023joint}                & 0.928    & 0.114    & 0.057    & 0.925      & 0.093     & 0.923           \\
                                                                                      & \citet{tonini2023objectaware}                 & 0.922    & 0.069    & 0.029    & 0.933      & 0.104     & 0.934           \\ \midrule
\multirow{2}{*}{\XSolid}                                                                     & \citet{tonini2023objectaware}$^\dagger$                  & 0.767    & 0.211    & 0.148    & -          & -         & -               \\
                                                                                      & \Name{} (ours)       & 0.966    & 0.080    & 0.029    & 0.972      & 0.064     & 0.951     \\ \bottomrule
\end{tabular}
\caption{Comparison between detection-based methods and \Name{}. Since none of the detection-based methods were evaluated on ChildPlay, we omit that dataset in this comparison. $^\dagger$Evaluation done by \citet{ryan2025gaze-lle} using \citet{tonini2023objectaware}'s public code.}
\label{tab:appendix-detection-comparison}
\end{table}

Despite using the conventional problem formulation and being systematically disadvantaged, \Name{} still outperforms detection-based methods in most metrics, underlining the strong performance of our method. When we remove ground-truth gaze matching, \Name{}'s advantage is even clearer. The results are in Table \ref{tab:appendix-detection-comparison}.

\end{document}